\documentclass[10pt,twocolumn,letterpaper]{article}

\usepackage[pagenumbers]{cvpr} %

\usepackage{microtype}

\renewcommand{\paragraph}[1]{\vspace{.5em}\noindent\textbf{#1.}}

\definecolor{cvprblue}{rgb}{0.21,0.49,0.74}
\usepackage[pagebackref,breaklinks,colorlinks,allcolors=cvprblue]{hyperref}

\usepackage{adjustbox}
\usepackage{glossaries}
\usepackage{glossaries-extra}
\usepackage{makecell}
\usepackage{multirow}
\usepackage{pgfplots}
\usepgfplotslibrary{groupplots}
\usepackage{tikz}
\usetikzlibrary{arrows.meta,shapes.geometric,calc,positioning}
\usepackage[dvipsnames]{xcolor}
\usepackage[table]{xcolor}
\usepackage{bbm}
\usepackage{array,tabularx}
\usepackage{tcolorbox}
\tcbuselibrary{breakable}
\usepackage{placeins}
\usepackage{gradient-text}
\usepackage{rotating}
\hypersetup{draft}

\usepackage{amsmath,amsfonts,bm}

\def\eqref#1{equation~\ref{#1}}

\def\1{\bm{1}}

\def\rvu{{\mathbf{i}}}

\def\rvu{{\mathbf{u}}}

\def\rvz{{\mathbf{z}}}

\def\rmP{{\mathbf{P}}}

\DeclareMathAlphabet{\mathsfit}{\encodingdefault}{\sfdefault}{m}{sl}
\SetMathAlphabet{\mathsfit}{bold}{\encodingdefault}{\sfdefault}{bx}{n}

\def\gA{{\mathcal{A}}}

\def\gC{{\mathcal{C}}}
\def\gD{{\mathcal{D}}}

\def\gF{{\mathcal{F}}}

\def\gI{{\mathcal{I}}}

\def\gM{{\mathcal{M}}}

\def\gT{{\mathcal{T}}}

\def\gY{{\mathcal{Y}}}
\def\gZ{{\mathcal{Z}}}

\def\sR{{\mathbb{R}}}

\newcommand{\softmax}{\mathrm{softmax}}

\newcommand{\normltwo}{L^2}

\DeclareMathOperator*{\argmax}{arg\,max}

\glsdisablehyper %
\makeglossaries

\newacronym{vit}{ViT}{Vision transformer}
\newacronym{resnet}{ResNet}{Residual neural network}
\newacronym{resnet18}{\textsc{ResNet18}}{ResNet-18}
\newacronym{resnet50}{\textsc{ResNet50}}{ResNet-50}
\newacronym{resnet101}{\textsc{ResNet101}}{ResNet-101}
\newacronym{mlp}{MLP}{Multi layer perceptron}
\newacronym{pp}{pp}{Percentage points}
\newacronym{ce}{CE}{Cross-entropy}
\newacronym{kl}{KL}{Kullback-Leibler divergence}
\newacronym{sgd}{SGD}{Stochastic gradient descent}
\newacronym{tta}{TTA}{Test-time augmentation}
\newacronym{ai}{AI}{Artificial intelligence}
\newacronym{dinov2}{\textsc{DINO}v2}{DINOv2: Learning Robust Visual Features without Supervision}
\newacronym{dinov3}{\textsc{DINO}v3}{DINOv3}

\newacronym{llm}{LLM}{Large language model}
\newacronym{vlm}{VLM}{Vision-language model}
\newacronym{lvlm}{LVLM}{Large vision-language model}
\newacronym{mllm}{MLLM}{Multimodal large language model}
\newacronym{fm}{FM}{Foundation model}
\newacronym{mfm}{MFM}{Multimodal foundation model}

\newacronym{clip}{\textsc{CLIP}}{Contrastive Language-Image Pre-training}
\newacronym{siglip}{\textsc{SigLIP}}{Sigmoid Loss for Language-Image Pre-training}
\newacronym{siglip2}{\textsc{SigLIP2}}{Multilingual Vision-Language Encoders with Improved Semantic Understanding, Localization, and Dense Features}
\newacronym{qwen}{\textsc{Qwen}}{Tongyi Qianwen family of LLMs}
\newacronym{qwen2}{\textsc{Qwen2}}{Tongyi Qianwen (2) family of LLMs}
\newacronym{qwen3}{\textsc{Qwen3}}{Tongyi Qianwen (3) family of LLMs}
\newacronym{qwen2vl2b}{\textsc{Qwen2-VL-2B}}{\textsc{Qwen2-VL-2B}}
\newacronym{qwen3vl2b}{\textsc{Qwen3-VL-2B}}{\textsc{Qwen3-VL-2B}}
\newacronym{qwen3vl8b}{\textsc{Qwen3-VL-8B}}{\textsc{Qwen3-VL-8B}}
\newacronym{qwen3vl32b}{\textsc{Qwen3-VL-32B}}{\textsc{Qwen3-VL-32B}}
\newacronym{cupl}{CuPL}{Customized prompts via language model}

\newacronym{domainnet}{DomainNet-126}{DomainNet-126}
\newacronym{oh}{Office-Home}{Office-Home}
\newacronym{visdac}{VisDA-C}{VisDA-C}
\newacronym{aircraft}{Aircraft}{Aircraft}
\newacronym{caltech}{Caltech}{Caltech101} 
\newacronym{car}{Car}{StandforCars} 
\newacronym{dtd}{DTD}{DTD} 
\newacronym{eurosat}{EuroSat}{EuroSat} 
\newacronym{flowers}{Flowers}{Flowers102} 
\newacronym{food}{Food}{Food101} 
\newacronym{imagenet}{ImageNet}{ImageNet} 
\newacronym{pets}{Pets}{OxfordPets} 
\newacronym{sun}{SUN}{SUN397} 
\newacronym{ucf}{UCF}{UCF101}

\newacronym{dnC}{C}{Clipart}
\newacronym{dnP}{P}{Painting}
\newacronym{dnR}{R}{Real}
\newacronym{dnS}{S}{Sketch}
\newacronym{ohar}{Ar}{Art}
\newacronym{ohcl}{Cl}{Clipart}
\newacronym{ohpr}{Pr}{Product}
\newacronym{ohrw}{Rw}{Real-World}
\newacronym{visdactrain}{Train}{Synthetic}
\newacronym{visdacvalidation}{Validation}{Real}

\newacronym{dl}{DL}{Deep learning}

\newacronym{uda}{UDA}{Unsupervised domain adaptation}
\newacronym{sfda}{SFDA}{Source-free domain adaptation}
\newacronym{sida}{SIDA}{Source-independent domain adaptation}
\newacronym{promd}{ProMD}{Proxy-manifold-guided denoising}

\usepackage{amsthm}
\theoremstyle{plain} 
\newtheorem{proposition}{Proposition}[section]
\newtheorem{corollary}{Corollary}[section]

\def\confName{CVPR}
\def\confYear{2027}

\title{%
Proxy-Manifold-Guided Denoising for Training-Free Adaptation of VLMs %
}

\author{%
Maximilian Forstenh\"ausler\textsuperscript{1,3},
Moritz Thoma\textsuperscript{2,3},
Emad Aghajanzadeh\textsuperscript{2,3}\\
Daniel K\"ulzer\textsuperscript{3},
Christos Anagnostopoulos\textsuperscript{1}\\[0.5em]
{\small\textsuperscript{1}University of Glasgow\quad\small\textsuperscript{2}Technical University of Munich\quad\small\textsuperscript{3}BMW Group}
}

\begin{document}
\addtocontents{toc}{\protect\setcounter{tocdepth}{2}}
\addtocontents{toc}{\protect\subsection*{Paper}}
\maketitle
\begingroup
\renewcommand{\thefootnote}{}
\footnotetext{\hspace*{-1.8em}\resizebox{\columnwidth}{!}{\scriptsize Preprint. Under review. Correspondence: m.forstenhaeusler.1@research.gla.ac.uk}}
\endgroup

\begin{abstract}
    In spite of its name, \glslink{sfda}{source-free domain adaptation (SFDA)}\glsunset{sfda} is not entirely source-free.\ Although source data remains inaccessible during adaptation, existing methods require a task-specific source-trained model.\ Recent \gls{sfda} approaches increasingly employ \glspl{vlm} to guide adaptation.\ %
    Despite outperforming conventional approaches, \gls{vlm}-guided methods exhibit limited variation in target-domain accuracy across various source-domain initializations.\ %
    This motivates improving the \gls{vlm} guidance and raises the question:\ can \gls{vlm} predictions that are refined independently of a task-specific source model provide more effective supervision for vision classifier adaptation?\ %
    We frame this perspective as \glslink{sida}{source-independent domain adaptation (SIDA)}\glsunset{sida}, a unifying taxonomy which advocates decoupling transductive prediction refinement of \glslink{fm}{foundation models (FMs)}\glsunset{fm} from downstream classifier training.\
    Building on this, we propose \glslink{promd}{\textbf{Pro}xy-\textbf{M}anifold-Guided \textbf{D}enoising (\textbf{ProMD})}\glsunset{promd}, a training-free adaptation approach for \glspl{vlm}.\ %
    \gls{promd} denoises zero-shot logits using nearest-neighbor consensus on a proxy manifold.\ 
    To address errors remaining after denoising occurred, \gls{promd} optionally queries a frozen \gls{lvlm} over a sample’s top-$5$ classes.\ 
    Across 14 datasets, \gls{promd} improves \gls{clip} ViT-B zero-shot accuracy by $\sim\kern-0.25em\mathbf{7}$\gls{pp}, while \gls{promd}${+}$ improves SOTA-refined accuracy by $\sim\kern-0.2em\mathbf{2.5}\text{--}\mathbf{3.5}$\gls{pp}, demonstrating benefits beyond zero-shot refinement. Within \gls{sida}, \gls{promd}'s pseudo-labels enable separately trained classifiers to outperform \gls{sfda} methods by $\sim\kern-0.2em\mathbf{3}$\gls{pp} without task-specific source training.

\end{abstract}

\section{Introduction}\label{sec:introduction}

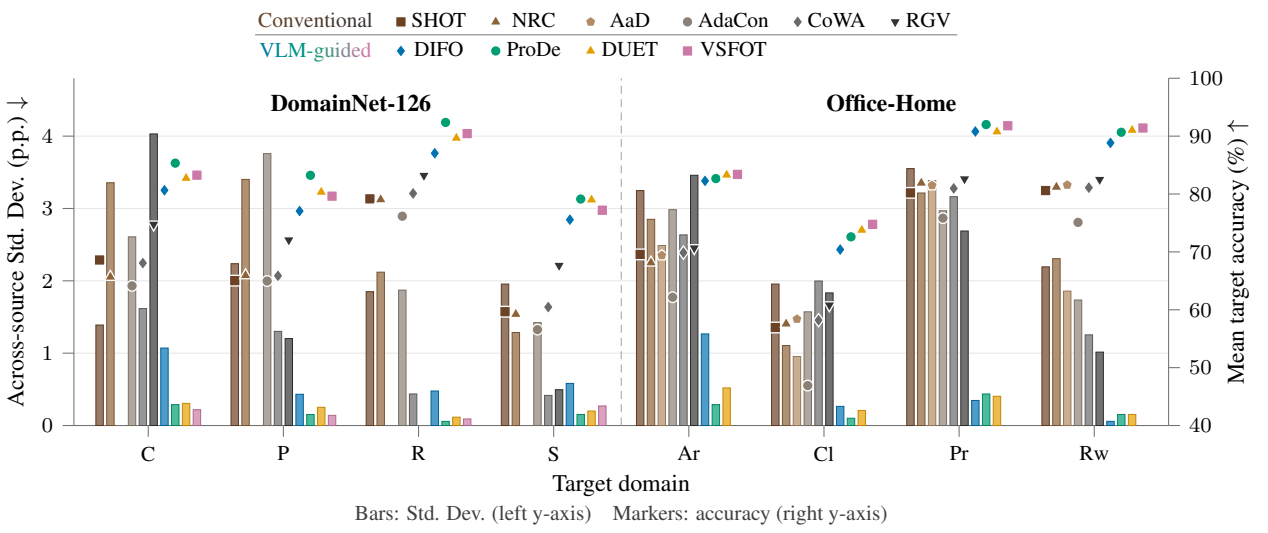
\begin{figure}[t]
    \centering
    \definecolor{cSHOT}{HTML}{6B4226}
\definecolor{cNRC}{HTML}{8C6239}
\definecolor{cAaD}{HTML}{B08B63}
\definecolor{cAdaCon}{HTML}{8C8175}
\definecolor{cCoWA}{HTML}{696969}
\definecolor{cRGV}{HTML}{343434}
\definecolor{cDIFO}{HTML}{0072B2}
\definecolor{cProDe}{HTML}{009E73}
\definecolor{cDUET}{HTML}{E69F00}
\definecolor{cVSFOT}{HTML}{CC79A7}
\pgfplotsset{
  sourcepaperaxis/.style={
    scale only axis, width=14.5cm, height=4.6cm,
    xmin=-0.55, xmax=7.55, enlarge x limits=false,
    axis line style={black!55,line width=0.4pt},
    tick align=outside, tick style={black!55,line width=0.4pt},
    font=\fontsize{9}{11}\selectfont,
    tick label style={font=\fontsize{8}{10}\selectfont},
    ylabel style={font=\fontsize{9}{11}\selectfont},
    scaled ticks=false,
  },
  sourcebar/.style={
    ybar, bar shift=0pt, bar width=2.8pt,
    draw=#1!85!black,fill=#1!70!white,line width=0.25pt,
  },
}
\adjustbox{max width=\linewidth}{%
\begin{tikzpicture}
\begin{axis}[
  sourcepaperaxis,name=stdaxis,
  axis x line*=bottom,axis y line*=left,
  ymin=0,ymax=4.8,ytick={0,1,2,3,4},
  ylabel={Across-source Std. Dev. (p.p.) $\downarrow$},
  ylabel style={at={(axis description cs:0.015,0.5)},anchor=north},
  xlabel={Target domain},
  xtick={0,...,7},xticklabels={C,P,R,S,Ar,Cl,Pr,Rw},
  ymajorgrids=true,grid style={black!10,line width=0.3pt},
]
\draw[black!35,densely dashed,line width=0.4pt]
  (axis cs:3.5,0)--(axis cs:3.5,4.8);
\node[anchor=north,font=\bfseries\fontsize{9}{11}\selectfont]
  at (axis cs:1.5,4.71) {DomainNet-126};
\node[anchor=north,font=\bfseries\fontsize{9}{11}\selectfont]
  at (axis cs:5.5,4.71) {Office-Home};
\addplot[sourcebar=cSHOT] coordinates {(-0.360,1.3892) (0.640,2.2368) (1.640,1.8502) (2.640,1.9553) (3.640,3.2470) (4.640,1.9553) (5.640,3.5511) (6.640,2.1932)};
\addplot[sourcebar=cNRC] coordinates {(-0.280,3.3561) (0.720,3.4020) (1.720,2.1197) (2.720,1.2858) (3.720,2.8501) (4.720,1.1060) (5.720,3.2140) (6.720,2.3065)};
\addplot[sourcebar=cAaD] coordinates {(3.800,2.4880) (4.800,0.9539) (5.800,3.3867) (6.800,1.8583)};
\addplot[sourcebar=cAdaCon] coordinates {(-0.120,2.6083) (0.880,3.7581) (1.880,1.8717) (2.880,1.4224) (3.880,2.9838) (4.880,1.5716) (5.880,2.9687) (6.880,1.7349)};
\addplot[sourcebar=cCoWA] coordinates {(-0.040,1.6166) (0.960,1.3013) (1.960,0.4359) (2.960,0.4163) (3.960,2.6350) (4.960,1.9975) (5.960,3.1628) (6.960,1.2530)};
\addplot[sourcebar=cRGV] coordinates {(0.040,4.0305) (1.040,1.2021) (3.040,0.4950) (4.040,3.4588) (5.040,1.8330) (6.040,2.6889) (7.040,1.0149)};
\addplot[sourcebar=cDIFO] coordinates {(0.120,1.0706) (1.120,0.4321) (2.120,0.4772) (3.120,0.5823) (4.120,1.2662) (5.120,0.2646) (6.120,0.3464) (7.120,0.0577)};
\addplot[sourcebar=cProDe] coordinates {(0.200,0.2887) (1.200,0.1528) (2.200,0.0577) (3.200,0.1528) (4.200,0.2887) (5.200,0.1000) (6.200,0.4359) (7.200,0.1528)};
\addplot[sourcebar=cDUET] coordinates {(0.280,0.3055) (1.280,0.2517) (2.280,0.1155) (3.280,0.2000) (4.280,0.5196) (5.280,0.2082) (6.280,0.4041) (7.280,0.1528)};
\addplot[sourcebar=cVSFOT] coordinates {(0.360,0.2183) (1.360,0.1405) (2.360,0.0907) (3.360,0.2702)};
\end{axis}
\begin{axis}[
  sourcepaperaxis,
  at={(stdaxis.south west)},anchor=south west,
  axis x line=none,axis y line*=right,
  ymin=40,ymax=100,ytick={40,50,60,70,80,90,100},
  xtick=\empty,ymajorgrids=false,
  ylabel={Mean target accuracy (\%) $\uparrow$},
  yticklabel pos=right,
  ylabel style={at={(axis description cs:1.13,0.5)},anchor=north},
  legend columns=6,legend cell align=center,
  legend style={
    at={(0.55,1.02)},anchor=south,draw=none,
    font=\fontsize{8}{10}\selectfont,
    /tikz/every even column/.append style={column sep=7pt},
  },
]
\addplot[only marks,mark=square*,mark size=2.0pt,
  mark options={solid,draw=white,fill=cSHOT,line width=0.5pt}]
  coordinates {(-0.360,68.6000) (0.640,65.0333) (1.640,79.1667) (2.640,59.6667) (3.640,69.5667) (4.640,56.9333) (5.640,80.2000) (6.640,80.6000)};
\addlegendentry{SHOT}
\addplot[only marks,mark=triangle*,mark size=2.5pt,
  mark options={solid,draw=white,fill=cNRC,line width=0.5pt}]
  coordinates {(-0.280,65.6667) (0.720,65.9333) (1.720,79.0333) (2.720,59.2333) (3.720,68.1333) (4.720,57.5667) (5.720,81.9000) (6.720,81.2000)};
\addlegendentry{NRC}
\addplot[only marks,mark=pentagon*,mark size=2.1pt,
  mark options={solid,draw=white,fill=cAaD,line width=0.5pt}]
  coordinates {(3.800,69.4000) (4.800,58.4000) (5.800,81.5000) (6.800,81.5667)};
\addlegendentry{AaD}
\addplot[only marks,mark=*,mark size=2.0pt,
  mark options={solid,draw=white,fill=cAdaCon,line width=0.5pt}]
  coordinates {(-0.120,64.1333) (0.880,64.9667) (1.880,76.1667) (2.880,56.5667) (3.880,62.1667) (4.880,46.9000) (5.880,75.8333) (6.880,75.1000)};
\addlegendentry{AdaCon}
\addplot[only marks,mark=diamond*,mark size=2.5pt,
  mark options={solid,draw=white,fill=cCoWA,line width=0.5pt}]
  coordinates {(-0.040,68.0667) (0.960,65.8667) (1.960,80.1000) (2.960,60.4667) (3.960,69.8667) (4.960,58.2000) (5.960,80.9667) (6.960,81.1000)};
\addlegendentry{CoWA}
\addplot[only marks,mark=triangle*,mark size=2.5pt,
  mark options={solid,draw=white,fill=cRGV,line width=0.5pt,rotate=180}]
  coordinates {(0.040,74.7500) (1.040,72.0500) (2.040,83.2000) (3.040,67.6500) (4.040,70.6667) (5.040,60.8000) (6.040,82.6000) (7.040,82.5000)};
\addlegendentry{RGV}
\addplot[only marks,mark=diamond*,mark size=2.5pt,
  mark options={solid,draw=white,fill=cDIFO,line width=0.5pt}]
  coordinates {(0.120,80.6500) (1.120,77.0600) (2.120,87.0433) (3.120,75.5633) (4.120,82.2667) (5.120,70.4000) (6.120,90.8000) (7.120,88.8333)};
\addlegendentry{DIFO}
\addplot[only marks,mark=*,mark size=2.0pt,
  mark options={solid,draw=white,fill=cProDe,line width=0.5pt}]
  coordinates {(0.200,85.3333) (1.200,83.2333) (2.200,92.3667) (3.200,79.1333) (4.200,82.6667) (5.200,72.6000) (6.200,92.0000) (7.200,90.6667)};
\addlegendentry{ProDe}
\addplot[only marks,mark=triangle*,mark size=2.5pt,
  mark options={solid,draw=white,fill=cDUET,line width=0.5pt}]
  coordinates {(0.280,82.7333) (1.280,80.3333) (2.280,89.6667) (3.280,79.0000) (4.280,83.3000) (5.280,73.7667) (6.280,90.7667) (7.280,91.0333)};
\addlegendentry{DUET}
\addplot[only marks,mark=square*,mark size=2.0pt,
  mark options={solid,draw=white,fill=cVSFOT,line width=0.5pt}]
  coordinates {(0.360,83.2567) (1.360,79.6267) (2.360,90.4367) (3.360,77.1967) (4.360,83.3900) (5.360,74.7600) (6.360,91.8000) (7.360,91.4000)};
\addlegendentry{VSFOT}
\end{axis}

\node[] (legend_title_a) at (3.2, 5.375) {\fontsize{8}{10}\selectfont \gradientRGB{Conventional}{107, 66, 38}{52, 52, 52}};
\node[] (legend_title_b) at (3.2, 4.95) {\fontsize{8}{10}\selectfont \gradientRGB{VLM-}{0, 114, 178}{0, 158, 115}\gradientRGB{guided}{0, 158, 115}{204, 121, 167}};
\draw[gray, top color=cSHOT, bottom color=cAaD, rounded corners=1pt] (2.4,5.175) -- (11.7,5.175);

\node[anchor=north,font=\fontsize{8}{10}\selectfont,text=black!75]
  at ([yshift=-0.93cm]stdaxis.south)
  {Bars: Std. Dev. (left y-axis)\quad Markers: accuracy (right y-axis)};
\end{tikzpicture}%
}
\vspace{-1.5em}
    \caption{Combined bar and scatter plot displaying source-domain sensitivity of target accuracy (std. dev., left y-axis) and mean target accuracy across source domains (right y-axis).}
    \label{fig:motivation-source-model-insensitivity}
\end{figure}

\glslink{dl}{Deep learning (DL)}\glsunset{dl} models can suffer substantial performance degradation under domain shift~\cite{quinonero-candela_dataset_2008}, i.e., when the deployment distribution differs from the training distribution.\ \glslink{uda}{Unsupervised domain adaptation (UDA)}\glsunset{uda} addresses this challenge by leveraging labeled source and unlabeled target data to improve target-domain performance. %
However, privacy, storage, and data-sharing constraints often prevent access to source data during adaptation~\cite{liang_we_2020}. %
\gls{sfda} proposes to instead adapt source-trained models using only unlabeled target data~\cite{liang_we_2020, yang_exploiting_2021, yang_attracting_2022}. Thus, while \gls{sfda} removes the need for source data, adaptation still relies on source-domain knowledge captured by the source-trained model.

Pretrained \glslink{vlm}{vision-language models (VLMs)}\glsunset{vlm} provide transferable knowledge acquired from large-scale image-text pretraining, enabling strong zero-shot classification~\cite{radford_learning_2021, zhai_sigmoid_2023, tschannen_siglip_2025}. 
Recent \gls{sfda} methods exploit this capability as guidance during the adaptation of source-trained models~\cite{tang_source-free_2024, tang_proxy_2025, lee_duet_2025, han_vision-language_2026}.\ Consistent with the observations of \citet{zhou_pros2t_2026}, investigating \gls{domainnet} and \gls{oh} (\cref{fig:motivation-source-model-insensitivity}) reveals that these methods achieve a higher mean target accuracy while exhibiting lower across-source standard deviation than conventional \gls{sfda} approaches~\cite{liang_we_2020, yang_exploiting_2021, yang_attracting_2022, chen_contrastive_2022, lee_confidence_2022, zhu_revisiting_2025}.\ %
Although limited sensitivity to source initialization does not establish that it is unnecessary, it motivates asking whether adaptation can proceed without it.\ \gls{vlm}-guided Fully-Source-Free Domain Adaptation (VF-SFDA)~\cite{zhou_pros2t_2026} explores this setting, with its Progressive Fusion with Structure--Shape Transformations (ProS$^2$T) method achieving competitive performance from random initialization.\
These findings motivate refining the supervision provided by \glspl{fm} in the target domain before transferring it to a downstream classifier.\ We formalize this perspective through \emph{\glslink{sida}{source-independent domain adaptation (SIDA)}\glsunset{sida}}.

In the absence of a task-specific source model, adaptation can draw on knowledge encoded in pretrained \glspl{vlm} and the structure of unlabeled target data.\ As shown in \cref{tab:sfda-benchmark-results}, zero-shot \gls{vlm} predictions already outperform conventional \gls{sfda} methods such as SHOT~\cite{liang_we_2020}, NRC~\cite{yang_exploiting_2021}, and AaD~\cite{yang_attracting_2022} on several evaluated domains, yet are exceeded by the evaluated \gls{vlm}-guided approaches~\cite{tang_source-free_2024, tang_proxy_2025, lee_duet_2025, han_vision-language_2026}. 
These results position zero-shot predictions as a strong baseline but leave room for target-domain refinement.\ Recent transductive refinement approaches~\cite{hu_sota_2026,
zanella_boosting_2024,li_efficient_2025} demonstrate that \gls{vlm} predictions can be adapted to target domains without task-specific source artifacts or updates to pretrained model parameters.
We hypothesize that image representations retain class-relevant information even when zero-shot \gls{vlm} predictions are incorrect. Standard zero-shot inference with \gls{clip}-based \glspl{vlm} classifies each image independently by comparing it with textual class descriptions, leaving relationships among target images unexploited.\ When these relationships reflect class structure, they can provide complementary evidence for refining uncertain or incorrect predictions~\cite{yang_exploiting_2021, yang_attracting_2022}. %
\definecolor{questionblue}{HTML}{27658E}
\begin{tcolorbox}[
  colback=questionblue!6!white,
  coltext=black,
  colframe=questionblue!6!white,
  boxrule=1pt,
  arc=0pt,
  outer arc=0pt,
  boxsep=0pt,
  left=4pt,
  right=4pt,
  top=4pt,
  bottom=4pt,
  before skip=6pt,
  after skip=6pt,
  width=\linewidth
]
We ask: \emph{how far can pretrained knowledge and unlabeled target data take prediction refinement and subsequent classifier training without task-specific source artifacts?}
\end{tcolorbox}
We propose \emph{\textbf{Pro}xy-\textbf{M}anifold-Guided \textbf{D}enoising} (\gls{promd}), a training-free approach that combines an ensemble of frozen \glspl{fm} to refine \gls{vlm} predictions using unlabeled target data.\ \gls{promd} comprises three stages: (i) constructing semantic predictions from a \gls{vlm}, (ii) denoising them using nearest-neighbor consensus from an independently selected visual representation, and (iii) selective reassessment of fragile predictions with a \glslink{lvlm}{large VLM (LVLM)}\glsunset{lvlm}.
All three stages use frozen pretrained \glspl{fm}. %
Our \textbf{contributions} are threefold:
\begin{itemize}
    \item We propose \textbf{\gls{promd}}, a training-free adaptation approach combining \textbf{proxy-manifold-guided denoising} with \textbf{selective semantic recovery} to refine zero-shot \gls{vlm} predictions or SOTA-refined predictions (\gls{promd}${+}$).
    \item We formulate \gls{sida}, a unifying transductive taxonomy without \textbf{task-specific source artifacts}, and advocate separating \gls{fm} adaptation from target-classifier training.
    \item We evaluate \gls{promd} directly against \textit{state-of-the-art} \gls{sfda} and \gls{vlm} adaptation methods across \textbf{14 datasets} and demonstrate its effectiveness for \emph{downstream distillation}.
\end{itemize}

\section{Related Work}\label{sec:related_work}
\textbf{Source-free domain adaptation (\gls{sfda})} adapts a source-trained model to an unlabeled target domain without accessing source data.\ SHOT~\cite{liang_we_2020} combines information maximization with pseudo-labeling while keeping the classifier frozen. NRC~\cite{yang_exploiting_2021} and AaD~\cite{yang_attracting_2022} exploit target neighborhoods, while AdaContrast~\cite{chen_contrastive_2022} combines contrastive learning with nearest-neighbor pseudo-label refinement. Other approaches use confidence-based sample weighting~\cite{lee_confidence_2022} or derive adaptation strategies from target-risk analysis~\cite{zhu_revisiting_2025}. More recent approaches incorporate diffusion models~\cite{huang_diffusion-driven_2025, wang_vicinity-guided_2025} into the adaptation process.

\usetikzlibrary{arrows.meta,calc,positioning}
\definecolor{sourcegold}{HTML}{C58A24}
\definecolor{targetblue}{HTML}{27658E}
\definecolor{vlmpurple}{HTML}{8056A3}
\definecolor{ink}{HTML}{27313B}
\definecolor{sidagreen}{HTML}{006864}
\tikzset{pics/frozen/.style={code={
      \foreach \a in {0,60,...,300}{
        \begin{scope}[rotate=\a, scale=1.1]
          \draw[targetblue!70,line width=.40pt,line cap=round]
            (0,0)--(.14,0) (.085,0)--(.115,.035)
            (.085,0)--(.115,-.035);
        \end{scope}
      }
    }}}

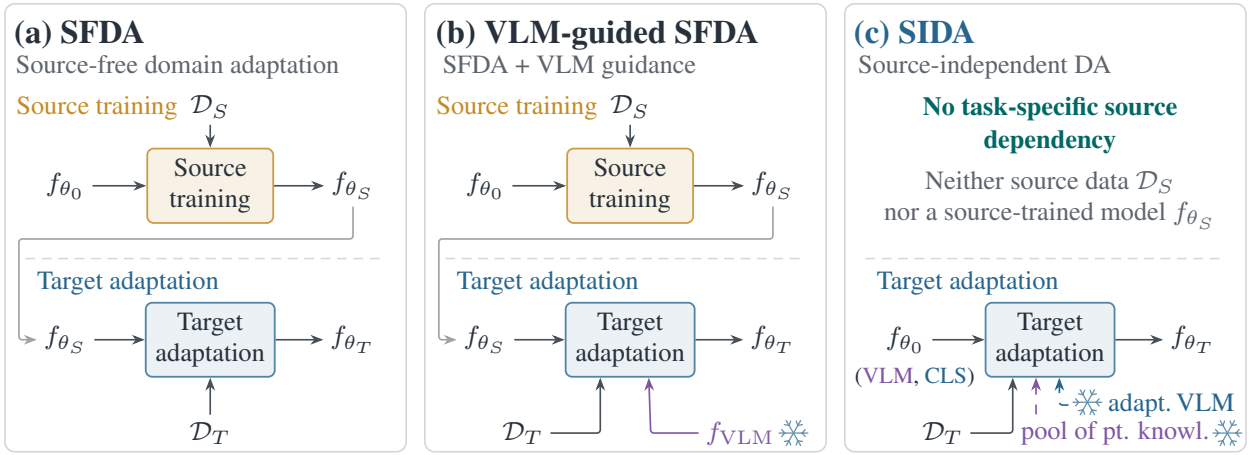
\begin{figure}
    \centering
\adjustbox{max width=\linewidth}{
\begin{tikzpicture}[
  x=1cm,y=1cm,
  font=\fontsize{8}{9.5}\selectfont,text=ink,
  >={Stealth[length=1.4mm,width=1.05mm]},
  flow/.style={->,draw=ink!85,line width=.5pt,rounded corners=1pt},
  guide/.style={flow,draw=vlmpurple},
  model/.style={inner sep=2pt},
  train/.style={draw=targetblue!80,fill=targetblue!9,
    rounded corners=2pt,line width=.55pt,align=center,
    minimum width=1.35cm,minimum height=.62cm,inner sep=3pt},
  source/.style={train,draw=sourcegold!85,fill=sourcegold!12},
  heading/.style={font=\bfseries\fontsize{10}{11.5}\selectfont},
  smallnote/.style={font=\fontsize{8}{9.5}\selectfont,text=ink!75},
  boundary/.style={draw=ink!20,densely dashed,line width=.45pt},
]
\foreach \x in {0,4.5,9}{
  \draw[draw=ink!20,fill=white,rounded corners=3pt,line width=.5pt]
    (\x,.90) rectangle ++(4.3,4.8);
  \draw[boundary] (\x+.22,3.22-0.25)--(\x+4.08,3.22-0.25);
}

\begin{scope}[name prefix=sfda-]
  \node[heading] at (0.8,5.62-0.25) {(a) SFDA};
  \node[smallnote] at (1.85,5.27-0.25) {Source-free domain adaptation};
  \node[smallnote,anchor=west,sourcegold] at (.0,4.83-0.25) {Source training};
  \node[model] (dataS) at (2.2,4.85-0.25) {$\mathcal D_S$};
  \node[model] (init) at (.65,4.00-0.25) {$f_{\theta_0}$};
  \node[source] (source) at (2.2,4.00-0.25) {Source\\training};
  \node[model] (sourceModel) at (3.73,4.00-0.25) {$f_{\theta_S}$};
  \draw[flow] (dataS)--(source);
  \draw[flow] (init)--(source);
  \draw[flow] (source)--(sourceModel);
  \node[smallnote,targetblue,anchor=west] at (.22,2.96-0.25) {Target adaptation};
  \node[model] (sourceInput) at (.65,2.35-0.25) {$f_{\theta_S}$};
  \node[train] (target) at (2.2,2.35-0.25) {Target\\adaptation};
  \node[model] (targetModel) at (3.73,2.35-0.25) {$f_{\theta_T}$};
  \node[model] (dataT) at (2.2,1.1) {$\mathcal D_T$};
  \draw[flow] (sourceInput)--(target);
  \draw[flow] (dataT)--(target);
  \draw[flow] (target)--(targetModel);
  \draw[flow,draw=ink!45] (sourceModel.south)--(3.73,3.43-0.25)
    --(.15,3.43-0.25)--(.15,2.35-0.25)--(sourceInput.west);
\end{scope}

\begin{scope}[xshift=4.5cm,name prefix=guided-]
  \node[heading] at (1.85,5.62-0.25) {(b) VLM-guided SFDA};
  \node[smallnote] at (1.55,5.27-0.25) {SFDA + VLM guidance};
  \node[smallnote,anchor=west,sourcegold] at (.0,4.83-0.25) {Source training};
  \node[model] (dataS) at (2.2,4.85-0.25) {$\mathcal D_S$};
  \node[model] (init) at (.65,4.00-0.25) {$f_{\theta_0}$};
  \node[source] (source) at (2.2,4.00-0.25) {Source\\training};
  \node[model] (sourceModel) at (3.73,4.00-0.25) {$f_{\theta_S}$};
  \draw[flow] (dataS)--(source);
  \draw[flow] (init)--(source);
  \draw[flow] (source)--(sourceModel);
  \node[smallnote,targetblue,anchor=west] at (.22,2.96-0.25) {Target adaptation};
  \node[model] (sourceInput) at (.65,2.35-0.25) {$f_{\theta_S}$};
  \node[train] (target) at (2.2,2.35-0.25) {Target\\adaptation};
  \node[model] (targetModel) at (3.73,2.35-0.25) {$f_{\theta_T}$};
  \node[model] (dataT) at (1.05,1.1) {$\mathcal D_T$};
  \pic at (3.95,1.1) {frozen};
  \node[model,text=vlmpurple] (vlm) at (3.38,1.1) {$f_{\mathrm{VLM}}$};
  \draw[flow] (sourceInput)--(target);
  \draw[flow] (dataT.east)--(1.88,1.1)--([xshift=-3.2mm]target.south);
  \draw[guide] (vlm.west)--(2.42,1.1)--([xshift=2mm]target.south);
  \draw[flow] (target)--(targetModel);
  \draw[flow,draw=ink!45] (sourceModel.south)--(3.73,3.43-0.25)
    --(.15,3.43-0.25)--(.15,2.35-0.25)--(sourceInput.west);
\end{scope}

\begin{scope}[xshift=9cm,name prefix=sida-]
  \node[heading,text=targetblue] at (0.75,5.62-0.25) {(c) SIDA};
  \node[smallnote] at (1.5,5.27-0.25) {Source-independent DA};
  \node[align=center,text=sidagreen,font=\bfseries\fontsize{8}{10}\selectfont]
    at (2.2,4.5-0.1) {\makecell[c]{No task-specific source\\dependency}};
  \node[smallnote,align=center] at (2.2,4-0.4)
    {Neither source data $\mathcal D_S$\\nor a source-trained model $f_{\theta_S}$};
  \node[smallnote,targetblue,anchor=west] at (.22,2.96-0.25) {Target adaptation};
  \node[model] (init) at (.65,2.35-0.25) {$f_{\theta_0}$};
  \node[train] (target) at (2.2,2.35-0.25) {Target\\adaptation};
  \node[model] (targetModel) at (3.73,2.35-0.25) {$f_{\theta_T}$};
  \draw[flow] (init)--(target);
  \draw[flow] (target)--(targetModel);
  \node[model] (dataT) at (1.05,1.1) {$\mathcal D_T$};
  \node[model,text=targetblue] (adpated_vlm) at (3.5,1.4) {adapt.\,VLM};
  \pic at (2.6,1.43) {frozen};
  \node[model,text=vlmpurple] (pretrained_knowledge) at (2.9,1.1) {pool of pt. knowl.};
  \pic at (4.1,1.1) {frozen};
  \node[model] (vlm) at (.75,1.7) {\scriptsize (\textcolor{vlmpurple}{VLM}, \textcolor{targetblue}{CLS})};
  \draw[flow] (dataT.east)--(1.8,1.1)--([xshift=-4mm]target.south);
  \draw[guide,targetblue,dashed] ([xshift=-3.5mm]adpated_vlm.west)--(2.3,1.4)--([xshift=1mm]target.south);
  \draw[guide, dashed] (2.05,1.3)--([xshift=-1.5mm]target.south);
\end{scope}
\end{tikzpicture}
}
    \caption{Comparison of \gls{sfda} paradigms: (a) Conventional \gls{sfda}, (b) \gls{vlm}-guided \gls{sfda}, and (c) the proposed \gls{sida}.
    }
    \label{fig:domain-adaptation-variants}
\end{figure}

\textbf{VLM-guided SFDA} uses semantic knowledge from pretrained \glspl{vlm} to guide adaptation of source-trained models. Most approaches build on \gls{clip}~\cite{radford_learning_2021}: DIFO~\cite{tang_source-free_2024} incorporates semantic knowledge through distillation; ProDe~\cite{tang_proxy_2025} explicitly corrects noisy \gls{vlm} predictions during adaptation; DUET~\cite{lee_duet_2025} combines agreement-based pseudo-label selection with consistency training; and VSFOT~\cite{han_vision-language_2026} uses VLM-derived semantic priors for optimal-transport alignment.\ However, \gls{vlm} guidance alone does not eliminate source-model dependence (cf. \cref{fig:domain-adaptation-variants}). ProS$^2$T~\cite{zhou_pros2t_2026} addresses this dependence by training a randomly initialized target model solely through \gls{vlm} guidance.

\textbf{Transductive refinement} exploits unlabeled target data to improve \gls{vlm} predictions.\ Related training-free methods exploit cached target
information (TDA~\cite{karmanov_efficient_2024} and zero-shot DMN~\cite{zhang_dual_2024}). InMaP~\cite{qian_intra-modal_2023} learns visual class proxies under textual guidance, while ZLaP~\cite{stojnic_label_2024} and ECALP~\cite{li_efficient_2025} refine predictions through
graph-based propagation. Training-free refinement can also involve auxiliary optimization while keeping pretrained models frozen: LAME~\cite{boudiaf_parameter-free_2022} optimizes a Laplacian-regularized objective; TransCLIP~\cite{zanella_boosting_2024} optimizes a transductive
objective; and SOTA~\cite{hu_sota_2026} integrates vision-only and vision-language \glspl{fm} through self-adaptive optimal transport. In contrast, GTA-CLIP~\cite{saha_generate_2025} alternates \gls{llm}-generated attribute expansion, transductive inference, and encoder fine-tuning.

Rather than iteratively optimizing prediction assignments, \gls{promd} interpolates semantic logits with neighborhood consensus from an independently selected frozen proxy manifold in a single step. An optional selective semantic recovery step then reassesses fragile predictions using a frozen \gls{lvlm}.

\section{Source-Independent Domain Adaptation}\label{sec:sida}

While \gls{sfda} removes access to task-specific source data \cite{liang_we_2020}, adaptation remains tied to a task-specific source-trained model.\
The limited sensitivity to task-specific source initialization observed in recent \gls{vlm}-guided \gls{sfda} methods questions its necessity, see \cref{fig:motivation-source-model-insensitivity}.\ Specifically, \citet{zhou_pros2t_2026} recently proposed the VF-SFDA setting and the ProS$^2$T method, which demonstrates that \gls{vlm} guidance can be competitive without source-specific initialization.\ These findings motivate centering adaptation on improving \glslink{fm}{foundation model (FM)}\glsunset{fm} supervision in the target domain, before transferring it to a task-specific classifier. %
We formulate \emph{\glslink{sida}{source-independent domain adaptation (SIDA)}\glsunset{sida}} as a unifying transductive umbrella taxonomy defined by its available resources (\cref{fig:domain-adaptation-variants}). \gls{sida} permits multiple transferable pretrained resources (pool of pt. knowledge) but excludes target labels and task-specific source artifacts (source data and source-trained classifiers) at adaptation time. The closed-set setting, considered in this work, assumes known target classes and their textual descriptions. Source independence concerns adaptation-time requirements and permits generic pretraining data to overlap with the target domain.\ %
Within this setting, we advocate separating \emph{target-domain prediction adaptation} from \emph{task-specific classifier training}. Predictions of \glspl{fm} may be adapted through \emph{transductive refinement}, which includes training-free or training-based approaches (incl. prompt or adapter training). The resulting predictions (adapt. VLM) can be used directly or distilled into a separate classifier. While the classifier does not require task-specific source training, we expect its backbone to be initialized from generic pretrained weights, e.g., \gls{imagenet}, to facilitate distillation. This taxonomy accommodates existing transductive zero-shot and training-free adaptation~\cite{saha_generate_2025, hu_sota_2026, li_efficient_2025, zhang_dual_2024} and VF-SFDA-style training~\cite{zhou_pros2t_2026}.

\section{Methodology}\label{sec:methodology}
\begin{figure*}
    \definecolor{mlInk}{HTML}{263445}
    \definecolor{mlMuted}{HTML}{647386}
    \definecolor{mlLine}{HTML}{D6DEE7}
    \definecolor{mlBlue}{HTML}{407DA7}
    \definecolor{mlPurple}{HTML}{8271A7}
    \definecolor{mlTeal}{HTML}{239C91}
    \definecolor{mlCoral}{HTML}{D97564}
    \definecolor{mlGold}{HTML}{D4AB4D}
    \tikzset{pics/frozen/.style={code={
      \foreach \a in {0,60,...,300}{
        \begin{scope}[rotate=\a, scale=1.1]
          \draw[mlBlue,line width=.60pt,line cap=round]
            (0,0)--(.14,0) (.085,0)--(.115,.035)
            (.085,0)--(.115,-.035);
        \end{scope}
      }
    }}}
    \centering
    \adjustbox{max width=0.96\textwidth}{%
    \begin{tikzpicture}[
      x=1cm,y=1cm,
      font=\sffamily\fontsize{8.5}{10}\selectfont,
      text=mlInk,
      >=Stealth,
      arr/.style={-{Stealth[length=1.7mm,width=1.2mm]},draw=mlMuted,line width=.65pt},
      mini/.style={font=\sffamily\fontsize{7.2}{8.5}\selectfont,text=mlMuted},
      lab/.style={font=\sffamily\fontsize{8}{9.5}\selectfont},
      box/.style={rounded corners=2pt,draw=mlLine,fill=white,line width=.65pt,align=center},
      dot/.style={circle,inner sep=0pt,minimum size=3.5pt},
      query/.style={diamond,inner sep=0pt,minimum size=6.5pt,draw=mlInk,line width=.85pt},
      selected/.style={circle,inner sep=0pt,minimum size=6.2pt,draw=mlInk,line width=.65pt}
    ]
    \path[use as bounding box] (0,.78) rectangle (18,7.4);

\filldraw[fill=mlBlue!2,draw=mlLine,rounded corners=3pt,line width=.7pt]
  (0,.80) rectangle (4.35,7.38);
\filldraw[fill=white,draw=mlLine,rounded corners=3pt,line width=.7pt]
  (4.65,.80) rectangle (13.35,7.38);
\filldraw[fill=mlMuted!1,draw=mlLine,rounded corners=3pt,line width=.7pt]
  (13.65,.80) rectangle (18,7.38);

\node[anchor=west,font=\sffamily\bfseries\fontsize{7.5}{9}\selectfont]
  at (.1,7.05) {(a)\enspace Zero-shot predictions};
\node[anchor=west,font=\sffamily\bfseries\fontsize{7.5}{9}\selectfont]
  at (4.7,7.05) {(b)\enspace Proxy-manifold-guided semantic denoising};
\node[anchor=west,font=\sffamily\bfseries\fontsize{7.5}{9}\selectfont]
  at (13.8,7.05) {(c)\enspace Constr. semantic recovery};
\draw[mlLine] (.15,6.8)--(4.2,6.8);
\draw[mlLine] (4.8,6.8)--(13.2,6.8);
\draw[mlLine] (13.8,6.8)--(17.8,6.8);

\node[box, line width=.35pt, rounded corners=1pt, inner sep=1pt, minimum width=1.26cm, minimum height=0.9cm] (img_box_3) at (1.05,5.9) {};
\node[box, line width=.35pt, rounded corners=1pt, inner sep=1pt, minimum width=1.26cm, minimum height=0.9cm] (img_box_2) at (1.0,5.95) {};
\node[box, line width=.35pt, rounded corners=1pt, inner sep=1pt] (img) at (0.95,6) {\includegraphics[width=1.2cm,height=0.9cm,keepaspectratio]{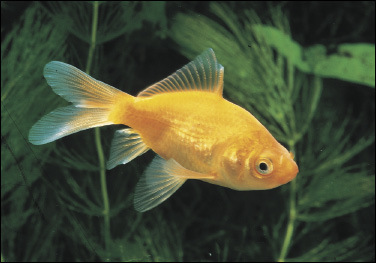}};
\foreach \id/\x/\dx/\t in {1/2/0.5/car,2/2.6/1.0/goldfish,3/3.7/0.5/$\dots$}{
  \draw[box, rounded corners=1pt] (\x, 6.3) rectangle (\x+\dx, 6.3+0.4);
  \node[font=\sffamily\bfseries\fontsize{6}{7.5}\selectfont] (class_\id) at (\x+\dx/2, 6.5) {\t};
}

\node[box,minimum width=1.55cm,minimum height=.82cm,lab,font=\sffamily\fontsize{7}{8.5}\selectfont] (prompt_box) at (3.1,5.65) {``a photo of\\\{article\} \{class\}''};

\node[box,fill=mlBlue!8,draw=mlBlue!45,minimum width=1.65cm,minimum height=.66cm, font=\sffamily\fontsize{7}{8.5}\selectfont]
  (imageenc) at (1.0,4.38) {Image\\[-1pt]Encoder $f_\theta$};
\node[box,fill=mlPurple!8,draw=mlPurple!45,minimum width=1.65cm,minimum height=.66cm, font=\sffamily\fontsize{7}{8.5}\selectfont]
  (textenc) at (3.1,4.38) {Text\\[-1pt]Encoder $g_\phi$};

\foreach \x in {1.625,3.725}{
    \pic at (\x,4.565) {frozen};
}
\node[mini] (image_emb) at (1.0,3.4) {$\mathbf{z}_{\mathcal{I},i}$};
\node[mini] (text_emb) at (3.1,3.4) {$\mathbf{z}_{\mathcal{T},c}$};
\node[mini] (cos_sim) at (2.0,3.05) {$\exp(\tau)\langle \,\cdot\,,\,\cdot\, \rangle$};

\draw[] (2.25,6.175)--(3.95,6.175);
\draw[] (2.25,6.3)--(2.25,6.175);
\draw[] (3.1,6.3)--(3.1,6.175);
\draw[] (3.95,6.3)--(3.95,6.175);
\draw[] (3.1, 6.175)--(3.1, 6.05);
\draw[arr] (img_box_2.south)--(imageenc.north);
\draw[arr] (prompt_box.south)--(textenc.north);
\draw[arr] (imageenc.south)--(image_emb.north);
\draw[arr] (textenc.south)--(text_emb.north);
\draw[arr] (1.3, 3.45)--(2.25, 3.45) -- (2.25, 3.2);
\draw[arr] (2.8, 3.45)--(2.565, 3.45) -- (2.565, 3.2);
\draw[arr] (2.05,2.9)--(2.05,2.6);

\node[font=\sffamily\bfseries\fontsize{7}{8.5}\selectfont] at (0.65,2.6) {\makecell[l]{Semantic\\proxy}};

\foreach \y/\a/\b/\c in {2.45-0.1/35/80/20,2.18-0.1/85/28/18,1.91-0.1/75/35/25,1.64-0.1/25/30/85}{
  \fill[mlTeal!\a] (1.34,\y) rectangle +(0.43,.21);
  \fill[mlCoral!\b] (1.81,\y) rectangle +(0.43,.21);
  \fill[mlGold!\c] (2.28,\y) rectangle +(0.43,.21);
}
\draw[mlInk,line width=.8pt] (1.27,2.70-0.1)--(1.20,2.70-0.1)--(1.20,1.59-0.1)--(1.27,1.59-0.1);
\draw[mlInk,line width=.8pt] (2.80,2.70-0.1)--(2.87,2.70-0.1)--(2.87,1.59-0.1)--(2.80,1.59-0.1);
\node at (2.09,1.14) {$\boldsymbol \Lambda^{(\mathrm{SP})}=\bigl[\ell_{i,c}^{(\mathrm{SP})}\bigr]_{i,c}\in\mathbb R^{N\times C}$};

\node[dashed, draw=mlGold, fill=mlGold!5, rounded corners=2pt, minimum width=1.35cm, minimum height=1cm] (promd_plus_box) at (4.2,2.7) {};
\node[mlMuted] (promd_plus) at (4.05,3) {\scriptsize\bfseries ProMD+};
\node[mlInk, draw=mlGold, fill=mlGold!5, rounded corners=2pt, minimum width=1.2cm, minimum height=0.4cm] (promd_plus) at (4.2,2.55) {};
\node[mlInk,] (promd_plus) at (4.2,2.55) {\scriptsize SOTA~\cite{hu_sota_2026}};
\draw[dashed, arr, mlGold] (3.35,1.85)--(3.35,2.7)--(promd_plus_box.west);
\draw[dashed, arr, mlGold] (promd_plus_box.east)--(5,2.7)--(5,1.85);
\node[mlMuted] (promd_plus) at (3.85,1.65) {\scriptsize\bfseries ProMD};

\draw[fill=mlMuted!1, draw=mlMuted!65, densely dashed,line width=.8pt, rounded corners=2pt] (8, 4.05) rectangle (13.05, 6.65);
\node[lab,font=\sffamily\fontsize{6.3}{7.3}\selectfont] at (11.5,3.9) {$\ell_2$-norm. visual feature space};
\node[lab,font=\sffamily\bfseries\fontsize{7}{8.5}\selectfont] at (6.25,6.4) {Semantic proxy $\left(\kern-0.2em\boldsymbol{\Lambda}^{(\mathrm{SP})}\kern-0.2em\right)$};
\node[lab,font=\sffamily\bfseries\fontsize{7}{8.5}\selectfont] at (9.1,6.4) {Proxy manifold};
\node[lab,font=\sffamily\bfseries\fontsize{7}{8.5}\selectfont] at (11.95,6.4) {Target manifold};
\draw[fill=mlMuted!3,draw=mlMuted!65,densely dashed,line width=.8pt] (6.15,5.2) circle (1.03);
\draw[fill=mlBlue!4,draw=mlBlue!75,line width=.9pt] (9.1,5.2) circle (1.03);
\draw[fill=mlPurple!4,draw=mlPurple!75,line width=.9pt] (11.95,5.2) circle (1.03);

\begin{scope}[shift={(6.15,4.81)}]
  \foreach \x/\y in {-.67/.04+.45,-.46/.46+.45,-.18/.46+.45,-.39/.18+.45,-.12/.08+.45}
    \node[dot,fill=mlTeal] at (\x,\y) {};
  \foreach \x/\y in {.25/.53+.45,.51/.30+.45,.65/.06+.45,.22/-.09+.45,.48/-.21+.45}
    \node[dot,fill=mlCoral] at (\x,\y) {};
  \foreach \x/\y in {-.56/-.47+.45,-.31/-.64+.45,-.02/-.54+.45,.25/-.50+.45,.47/-.58+.45}
    \node[dot,fill=mlGold] at (\x,\y) {};
  \foreach \x/\y in {-.46/.46+.45,-.18/.46+.45,-.39/.18+.45,.22/-.09+.45}
    \node[selected] at (\x,\y) {};
  \node[query,fill=mlCoral] at (.38,.10+.45) {};
\end{scope}

\begin{scope}[shift={(9.1,4.81+.42)}]
  \draw[mlBlue!50,densely dashed,line width=.7pt,fill=mlBlue!5] (-.27,.18) ellipse (.46 and .42);
  \foreach \x/\y in {-.58/.23,-.38/.48,-.13/.43,-.07/-.03}
    \draw[mlBlue!55,line width=.55pt] (-.27,.18)--(\x,\y);
  \foreach \x/\y in {.25/.52,.49/.42,.70/.12,.50/-.14,.25/-.36,-.18/-.61,-.55/-.45,-.78/-.20,.57/-.67}
    \node[dot,fill=mlMuted!65] at (\x,\y) {};
  \foreach \x/\y in {-.58/.23,-.38/.48,-.13/.43,-.07/-.03}{
    \node[dot,fill=mlBlue] at (\x,\y) {};
    \node[selected] at (\x,\y) {};
  }
  \node[query,fill=white] at (-.27,.18) {};
  \node[mini,text=mlBlue, font=\sffamily\fontsize{6.5}{8}\selectfont] at (-0.0,.725) {$\mathcal N_{m,k}(i)$};
\end{scope}

\begin{scope}[shift={(11.95,4.81)}]
  \draw[mlTeal!20,line width=9pt,line cap=round] (-.66,.02+.45)..controls(-.47,.66+.45)and(-.04,.43+.45)..(-.13,.15+.45);
  \draw[mlCoral!17,line width=9pt,line cap=round] (.21,.44+.45)..controls(.66,.34+.45)and(.78,-.02+.45)..(.48,-.26+.45);
  \draw[mlGold!20,line width=9pt,line cap=round] (-.56,-.49+.45)..controls(-.20,-.73+.45)and(.17,-.66+.45)..(.26,-.42+.45);
  \foreach \x/\y in {-.63/.07+.45,-.50/.31+.45,-.43/.49+.45,-.21/.39+.45,-.11/.17+.45,-.40/.08+.45}
    \node[dot,fill=mlTeal] at (\x,\y) {};
  \foreach \x/\y in {.24/.43+.45,.49/.40+.45,.66/.22+.45,.59/-.03+.45,.44/-.22+.45}
    \node[dot,fill=mlCoral] at (\x,\y) {};
  \foreach \x/\y in {-.52/-.46+.45,-.34/-.62+.45,-.10/-.62+.45,.10/-.53+.45,.24/-.40+.45}
    \node[dot,fill=mlGold] at (\x,\y) {};
  \node[query,fill=mlTeal] at (-.25,.17+.45) {};
\end{scope}

\draw[arr] (7.90,4.81+.45)--(7.25,4.81+.45);
\node[mini,align=center,font=\sffamily\fontsize{6.3}{7.3}\selectfont] at (7.57,5.17+.45) {retrieve\\scores};
\draw[draw=mlMuted!75,densely dashed,line width=.7pt] (10.20,4.81++.45)--(10.84,4.81+.45);
\node[mini,align=center,font=\sffamily\fontsize{6.3}{7.3}\selectfont] at (10.52,5.23+.25) {approx.};
\draw[arr] (6.15,4.1)--(6.15,3.85)--(9,3.85)--(9,3.5);
\node[mini,inner sep=1pt,font=\sffamily\fontsize{7}{8.5}\selectfont] at (7, 4) {manifold};
\node[mini,inner sep=1pt,font=\sffamily\fontsize{7}{8.5}\selectfont] at (7, 3.7) {consensus};

\foreach \cx/\title/\sym in {5.85/Original scores/{\boldsymbol\ell_i},8.7/Neighbor consensus/{\bar{\boldsymbol\ell}^{(m,k)}_i},11.55/Refined scores/{\widetilde{\boldsymbol\ell}^{(m,k)}_i}}{
  \node[lab] at (\cx+0.5,3.2) {$\sym$};
  \draw[mlMuted!70,line width=.55pt] (\cx-.2,1.03+.45)--(\cx-.2,2.37+.45);
  \draw[mlMuted!70,line width=.55pt] (\cx-.2,1.03+.45)--(\cx+1.3,1.03+.45);
  \draw[mlLine,densely dotted] (\cx-.2,1.68+.45)--(\cx+1.3,1.68+.45);
  \draw[mlLine,densely dotted] (\cx-.2,2.33+.45)--(\cx+1.3,2.33+.45);
  \foreach \dx/\lab in {0.15/1,.55/2,.95/3}
    \node[mini] at (\cx+\dx,.86+.45) {$\lab$};
}
\foreach \cx/\a/\b/\c in {5.85/.85/1.20/.40,8.7/1.40/.60/.25,11.55/1.125/.9/.325}{
  \fill[mlTeal] (\cx,1.04+.45) rectangle (\cx+.3,1.04+\a+.45);
  \fill[mlCoral] (\cx+.4,1.04+.45) rectangle (\cx+.7,1.04+\b+.45);
  \fill[mlGold] (\cx+.8,1.04+.45) rectangle (\cx+1.1,1.04+\c+.45);
}
\node[font=\fontsize{14}{14}\selectfont,text=mlMuted] at (7.8,2.4) {$+$};
\draw[arr] (10.3, 2.3)--(11,2.3);
\node[mini] at (7.8, 2.1) {blend};
\draw[arr] (3.2, 1.85)--(5.35, 1.85);

\begin{scope}[font=\sffamily\fontsize{6.6}{7.6}\selectfont]
\draw[arr] (13.05,2.3)--(13.48,2.3)--(13.48,6.28)--(14.19,6.28);
\node[box,draw=mlCoral!65,fill=mlCoral!7, minimum width=3.54cm,minimum height=.55cm,inner sep=3pt] (fragile) at (15.95,6.38) {\textbf{Fragile subset $\mathcal F$}%
};
\node[box,draw=mlGold!75,fill=mlGold!9, minimum width=3.54cm,minimum height=.68cm,inner sep=3pt](priority) at (15.95,5.45)
 {\textbf{Proxy-error ranking}\\[1pt]
  selects top $(1{-}q)$ $\longrightarrow\mathcal F_q \subseteq \mathcal{F}$};
\draw[arr] (fragile.south)--(priority.north);

\draw[draw=mlPurple,line width=.65pt] (priority.south)--(15.95,4.85);

\draw[arr,draw=mlPurple] (15.95,4.85)--(14.78,4.85)--(14.78,4.55);
\draw[arr,draw=mlPurple] (15.95,4.85)--(16.67,4.85)--(16.67,4.60);
\node[box,draw=mlBlue!40,inner sep=2pt] (rec_image) at (14.78,4.15)
 {\includegraphics[width=.91cm,height=.68cm,keepaspectratio]
  {static/images/n01443537_goldfish.JPEG}};
\node[box,draw=mlPurple!45,fill=mlPurple!4,
 minimum width=2.10cm,minimum height=.88cm,inner sep=2pt]
 (candidate_ids) at (16.65,4.15)
 {\textbf{Top-$k_\gC$ candidates $\mathcal C_i$}\\[1pt]
  \hspace{2em}\makecell[l]{\texttt{C00}: car \\
  \texttt{C01}: goldfish\\$\cdots$}};

\node[box,draw=mlPurple!65,fill=mlPurple!10,
 minimum width=2.8cm, minimum height=1.05cm,inner sep=3pt]
 (qwen_score) at (15.95,2.75)
 {\textbf{\gls{qwen3vl2b}}\\[1pt]
  Top-$k_\gC$ candidate\\[1pt]constrained ID scoring};
\draw[arr, draw=mlPurple,] (rec_image.south)--(14.78,3.275);
\draw[arr, draw=mlPurple,] (candidate_ids.south)--(16.65,3.275);
\pic at (17.1,3.05) {frozen};
\draw[arr,draw=mlPurple] (qwen_score.south)--(15.95,1.95);

\node[box,draw=mlTeal!75,fill=mlTeal!15,
 minimum width=3.52cm,minimum height=.40cm,inner sep=2pt]
 (final_label) at (15.95,1.15) {\textbf{Final pseudo-label} $\widehat{y}_i$};
 
\draw[arr,draw=mlTeal] (15.95,1.65)--(final_label.north);
\draw[arr,draw=mlMuted!85,rounded corners=1.5pt]
 (priority.west)--(13.92,5.45)--(13.92,1.8)--(15.95-0.15,1.8);
\node[rotate=90,fill=mlMuted!1,inner sep=2pt,
 font=\sffamily\fontsize{6}{7}\selectfont]
 at (13.92,3.30) {$i\notin\mathcal F_q$: retain $\hat y_i^{(\mathrm D)}$};
 \draw[] (15.95,1.8) circle (0.15);
 \draw[] (15.95,1.65) -- (15.95, 1.95);
 \draw[] (15.95-0.15,1.8) -- (15.95+0.15, 1.8);
\end{scope}
\end{tikzpicture}%
    }
    \caption{The three stages of \gls{promd}: (a) displays the creation of the proxy-semantic space $\boldsymbol{\Lambda}^{(\mathrm{SP})}$, (b) illustrates the proxy-manifold-guided denoising scheme, and (c) shows the constrained semantic recovery. \protect\tikz[baseline=-0.5ex, scale=0.9]{
      \protect\pic {frozen};
    } denotes frozen model weights.
    }
    \label{fig:method}
\end{figure*}

\paragraph{Notation}
We follow the notation of \citet{goodfellow_deep_2016}. Let $\gI\subseteq \sR^{H\times W\times C_{\mathrm{img}}}$ denote the space of preprocessed images and $\gT$ the space of tokenized text sequences.\ A text sequence is written as $t{=}(a_l)_{l=1}^L\in\gT$, where $a_l\in\gA$ is a token from vocabulary $\gA$.\ We denote the image and text encoders by $f_\theta$ and $g_{\phi}$, respectively. Unless stated otherwise, these encoders include their pretrained projection heads, which map the respective representations into a shared embedding space $\gZ\subseteq\sR^{d}$. We define $\normltwo$-normalization as $\nu(\mathbf{z})=\mathbf{z}/\|\mathbf{z}\|_2$ for $\mathbf{z}\neq\mathbf{0}$.

\paragraph{Problem formulation} 
Let $\gD_T{=}\{x_i\}_{i=1}^{N}$ be an unlabeled target dataset with a known set of $C$ classes.\ Let $T{=}(t_c)_{c=1}^C$ collect their tokenized class descriptions.\ A pretrained frozen \gls{vlm} induces a zero-shot scoring function $h_{\mathrm{VLM}}: \gI\times\gT^{C}\rightarrow\sR^{C}$, yielding $\boldsymbol{\ell}_i{=}h_{\mathrm{VLM}}(x_i,T)$ and %
$\hat y_i{=}\arg\max_{c\in\left[C\right]}\ell_{i,c}$. Under \gls{sida}, let $\Psi$ denote a pool of pretrained models.\ Our objective is to refine zero-shot logits, %
\begin{equation}
    \widetilde{\boldsymbol{\Lambda}}
    = \Phi(\mathcal{D}_T,T;\Psi)
    = \bigl[\widetilde{\ell}_{i,c}\bigr]_{i,c}
    \in \sR^{N\times C},
\end{equation}
to improve their target classification accuracy over the zero-shot counterparts.\ The operator $\Phi$ comprises zero-shot scoring and target-domain adaptation, which may involve model training.\ For \gls{promd}, all pretrained models of $\Psi$ remain frozen.\ Under \gls{sida}, the resulting pseudo-labels or class-distributions can then be used as supervision for additional classifier training in the target domain.

\paragraph{Overview} As illustrated in \cref{fig:method}, \gls{promd} comprises three stages: First, \gls{promd} constructs a \emph{proxy-semantic space} from the \gls{vlm}'s zero-shot logits (class scores) (\cref{sec:method-semantic-proxy}). Second, \gls{promd} gathers the $k$ nearest neighbors from visual representations on a \emph{proxy manifold} to form a consensus in the proxy-semantic space, which it uses to \emph{denoise} the initial predictions (\cref{sec:method-proxy-denoising}). Third, we identify fragile predictions and apply \emph{semantic recovery} through a \gls{lvlm}, using constrained inputs to guide reassessment (\cref{sec:method-semantic-recovery}). Together, the three stages combine pretrained semantic knowledge, target-data geometry, and visual-language reasoning to refine target predictions training-free.

\subsection{Zero-Shot Semantic Proxy Construction}\label{sec:method-semantic-proxy}
Following prior work~\cite{tang_source-free_2024, tang_proxy_2025, lee_duet_2025, han_vision-language_2026, stojnic_label_2024, li_efficient_2025}, 
we use \gls{clip}~\cite{radford_learning_2021} to obtain zero-shot predictions that serve as an initial semantic proxy for $\gD_T$.\ While informative, these predictions are noisy under domain shift (cf. \cref{tab:sfda-benchmark-results} and \ref{tab:categorization-benchmark-results}), motivating their refinement.\ Let $f_\theta$ and $g_\phi$ denote the frozen image and text encoders, respectively, and $\exp(\tau)$ a pretrained log inverse-temperature scale. For a target image $x_i \in \gI$ and a tokenized class description $t_c$, the zero-shot logit is given by
\begin{equation}
    \ell_{i,c}^{(\mathrm{SP})}
    =
    \exp(\tau)
    \left\langle
        \nu(f_\theta(x_i)),
        \nu(g_\phi(t_c))
    \right\rangle,
    \label{eq:vlm-logits}
\end{equation}
where $\tau$ is fixed throughout adaptation.\ The respective probability and prediction are denoted by $p^{(\mathrm{SP})}_{i,c}=\softmax(\ell_{i,c}^{(\mathrm{SP})})$, and $
    \hat y_i^{(\mathrm{SP})}
    {=}
    \arg\max_{c\in[C]}
    p^{(\mathrm{SP})}_{i,c}$. Stacking across target samples yields
the semantic proxy representation:
\begin{equation}
    \boldsymbol{\Lambda}^{(\mathrm{SP})}
    =
    \bigl[\ell_{i,c}^{(\mathrm{SP})}\bigr]_{i,c},
    \qquad
    \mathbf{P}^{(\mathrm{SP})}
    =
    \bigl[p^{(\mathrm{SP})}_{i,c}\bigr]_{i,c},
    \label{eq:semantic-proxy}
\end{equation}
where $\boldsymbol{\Lambda}^{(\mathrm{SP})}, \mathbf{P}^{(\mathrm{SP})}{\in}\sR^{N\times C}$.  %
Beyond zero-shot predictions, we also consider semantic proxies from already-refined predictions. We introduce \textbf{\gls{promd}${\mathbf{+}}$}, which uses SOTA's~\cite{hu_sota_2026} transductively refined predictions for subsequent proxy-manifold denoising and semantic recovery.\ Given $p_{i,c}^{(\mathrm{SP},\mathrm{SOTA})}$ for sample $i$ and class $c$, we apply \gls{promd} in log-probability space (clamped to ensure finite scores)
\begin{equation}
    \ell_{i,c}^{(\mathrm{SP},\mathrm{SOTA})} = \log (\max (p_{i,c}^{(\mathrm{SP},\mathrm{SOTA})}, 10^{-30})).
    \label{eq:sota-log-prob-semantic-proxy}
\end{equation}

\subsection{Proxy-Manifold-Guided Semantic Denoising}\label{sec:method-proxy-denoising}
\paragraph{Noisy prediction model} Pretrained \glspl{vlm} provide imperfect supervision under target-domain shift \cite{tang_proxy_2025}. Their predictions can guide adaptation, but treating them as reliable pseudo-labels risks propagating domain-specific errors. We formalize this uncertainty as,
\begin{equation}
    \boldsymbol{\ell}_i
    =
    \rvu_i+\boldsymbol{\varepsilon}_i, \qquad \boldsymbol{\ell}_i, \rvu_i, \boldsymbol{\varepsilon}_i \in \sR^C.
    \label{eq:noisy-prediction-model}
\end{equation}
For analysis, the \emph{noisy prediction model} decomposes the observed logit $\boldsymbol{\ell}_i$ into a clean, unobserved logit vector $\rvu_i$ and noise $\boldsymbol{\varepsilon}_i$, which captures deviations associated with domain shift or prompt mismatch. %

\paragraph{Neighborhood construction and logit denoising} 
\gls{promd} builds on the premise that semantically related target samples offer complementary information that can improve individual predictions.\ 
The visual representation used to identify these relationships may come from a different encoder than the one producing the semantic proxy; a proxy-target geometry determines neighborhoods while the \gls{vlm} provides class semantics.
Let $m$ index a frozen visual encoder $f_{\theta}^{(m)}$, and let $\rvz^{(m)}_i{=}\nu(f_{\theta}^{(m)}(x_i))$ denote its normalized representation of $x_i$. We construct the self-excluding neighborhood
\begin{equation}
    \mathcal{N}_{m,k}(i)
    =
    \operatorname*{TopK}_{j\in[N]\setminus\{i\}}
    \left\langle
        \nu(f_{\theta}^{(m)}(x_i)), \nu(f_{\theta}^{(m)}(x_j))
    \right\rangle,
    \label{eq:proxy-neighbors}
\end{equation}
where $1{\leq}k{<}N$, $[N]{=}\{1,\ldots,N\}$ and $\operatorname{TopK}$ returns the indices of the $k$ nearest neighbors quantified by cosine similarity. We then aggregate the semantic consensus based on the visual similarity in the proxy manifold $m$, and interpolate to produce a denoised logit score as
\begin{equation}
    \widetilde{\boldsymbol{\ell}}^{(m,k)}_i
    =
    (1-\lambda_{\mathrm{den}})\boldsymbol{\ell}_i
    +
    \lambda_{\mathrm{den}} \bar{\boldsymbol{\ell}}_i^{(m,k)},
    \label{eq:denoised-logit}
\end{equation}
where $\bar{\boldsymbol{\ell}}_i^{(m,k)} {=}\frac1k \sum_{j\in\mathcal{N}_{m,k}(i)}\boldsymbol{\ell}_j$ represents the averaged manifold consensus, $k$ controls the neighborhood size, and $\lambda_{\mathrm{den}}\kern-0.2em\in\kern-0.2em[0,1]$ its contribution.\ Visual similarity in manifold $m$ determines which samples exchange evidence, while the semantic proxy supplies their class scores.\ The interpolation performs a single smoothing step using the original scores and the manifold consensus. %

\paragraph{Denoising rationale - theoretical perspective}
We investigate the conditions under which interpolating a \gls{vlm} logit vector with proxy-neighborhood consensus yields lower error relative to its unobserved clean logit.\ 
Given \cref{eq:noisy-prediction-model}, we condition all expectations on the unobserved clean logits and the realized proxy neighborhood, and require only finite second moments; neither unbiasedness nor independence is assumed. For a target sample $x_i$ and proxy manifold $m$, subtracting $\mathbf{u}_i$ from $\widetilde{\boldsymbol{\ell}}_i^{(m,k)}$ in \cref{eq:denoised-logit} yields

\begin{equation}
    \widetilde{\boldsymbol{\ell}}_i^{(m,k)}
    -
    \mathbf{u}_i
    =
    (1-\lambda_{\mathrm{den}})\boldsymbol{\varepsilon}_i
    +
    \lambda_{\mathrm{den}}\boldsymbol{\eta}_i,
\label{eq:denoised-error}
\end{equation}
where $\boldsymbol{\eta}_i$ is the error of the retrieved neighbor consensus,
\begin{equation}
    \boldsymbol{\eta}_i
    =
    \bar{\boldsymbol{\ell}}_i^{(m,k)}
    -
    \mathbf{u}_i
    =
    \mathbf{b}_i
    +
    \frac{1}{k}
    \sum_{j\in\mathcal{N}_{m,k}(i)}
    \boldsymbol{\varepsilon}_j,
\label{eq:theory-consensus-error}
\end{equation}
and
    $\mathbf{b}_i
    {=}
    \frac{1}{k}
    \sum_{j\in\mathcal{N}_{m,k}(i)}
    \mathbf{u}_j
    -
    \mathbf{u}_i$
captures the semantic mismatch introduced by neighborhood aggregation.

\begin{proposition}[Risk of neighborhood interpolation]\label{prop:risk-neighborhood-interpolation}
Let
\begin{align*}
 &r_i{=}\mathbb{E} \bigl[ \|\boldsymbol{\varepsilon}_i\|_2^2\bigr], \qquad s_i{=}\mathbb{E}\bigl[\|\boldsymbol{\eta}_i\|_2^2\bigr],\\
 &\chi_i{=}\mathbb{E}\bigl[\langle\boldsymbol{\varepsilon}_i,\boldsymbol{\eta}_i\rangle\bigr], \quad d_i {=}r_i+s_i-2\chi_i,
\end{align*}
where $r_i$ and $s_i$ denote the expected squared error of the semantic proxy prediction and the retrieved neighbor consensus, respectively, $\chi_i$ denotes their expected error alignment, and $d_i$ denotes their expected squared discrepancy. Given $d_i{=}\mathbb{E}\bigl[\|\boldsymbol{\eta}_i-\boldsymbol{\varepsilon}_i\|_2^2\bigr]\geq0$, the risk is
\begin{equation}
    R_i(\lambda_{\mathrm{den}})
    {=}
    \mathbb{E}\kern-0.25em
    \left[\kern-0.15em\left\|
        \widetilde{\boldsymbol{\ell}}_i^{(m,k)}\kern-0.1em{-}\mathbf{u}_i
    \right\|_2^2\right]\kern-0.25em
    {=}
    r_i-2\lambda_{\mathrm{den}}(r_i-\chi_i)+\lambda_{\mathrm{den}}^2d_i.
\label{eq:exact-denoising-risk}
\end{equation}
For $d_i>0$ and $\lambda_{\mathrm{den}}\in(0,1]$, $R_i(\lambda_{\mathrm{den}})<r_i$ iff
\begin{equation}
    \lambda_{\mathrm{den}}<
    2(r_i-\chi_i) / d_i.
\label{eq:exact-improvement-condition}
\end{equation}
If $d_i=0$, then $R_i(\lambda_{\mathrm{den}})=r_i$ for every $\lambda_{\mathrm{den}}$.
\end{proposition}

By \cref{prop:risk-neighborhood-interpolation}, interpolation strictly reduces risk \emph{if and only if} the expected error alignment between $\boldsymbol{\ell}_i$ and $\bar{\boldsymbol{\ell}}_i^{(m,k)}$ is smaller than the expected squared error of $\boldsymbol{\ell}_i$ ($\chi_i{<}r_i$) and $\lambda_{\mathrm{den}}$ satisfies \cref{eq:exact-improvement-condition}. Intuitively, $\chi_i{<}r_i$ states that denoising must, on average, reduce the component of error along $\boldsymbol{\ell}_i$'s error direction. Interpolation then partially corrects the original error, while a sufficiently small $\lambda_{\mathrm{den}}$ limits the influence of other errors introduced by the consensus. This explains why improvement remains possible even when neighbor-consensus may have higher individual risk than the original $(s_i{>}r_i)$. Since $\boldsymbol{\eta}_i$ includes both neighborhood semantic mismatch and neighboring prediction errors, denoising improvement is not guaranteed unconditionally. %

\subsubsection{Proxy-Manifold Selection}\label{sec:method-proxy-manifold-selection}
Since $\mathcal{D}_T$ is unlabeled and the proxy-manifold $m$ is independent of the \gls{vlm} backbone in \cref{eq:denoised-logit}, a ground-truth-free selection strategy is required to identify the most suitable manifold for refinement in $\mathcal{D}_T$.\  
From a fixed set $\gM {=} \{m_j\}_{j=1}^{J=4}$ of proxy manifolds constructed from features extracted by \gls{clip}~\cite{radford_learning_2021}, \gls{dinov2}~\cite{oquab_dinov2_2024}, \gls{siglip2}~\cite{tschannen_siglip_2025}, and \gls{qwen3vl2b}~\cite{bai_qwen3-vl_2025}, we seek the manifold that best supports prediction refinement.\
We choose the \emph{agreement} of proxy-semantic predictions across candidate manifolds as the selection criterion.\  
For each candidate $m$, let $\mathbf{Z}^{(m)}{=}[\nu(f_{\theta}^{(m)}(x_i))^\top]_{i=1}^{N}\in\sR^{N\times d_m}$ denote the normalized visual features.\ Through the construction of a self-excluding $k$-nearest-neighbor graph we obtain the manifold-specific predictions $\hat y_i^{(m,k,w)}
{=}
\arg\max_{c\in[C]}p^{(m,k,w)}_{i,c}$, by applying a fixed propagation to the frozen semantic proxy probabilities $\rmP^{(m,k,w)}{=}\operatorname*{Prop}_{k,w} \left(\mathbf{P}^{(\mathrm{SP})};\mathbf{Z}^{(m)}\right)$, %
where $w$ denotes the neighbor weighting rule.
To prevent each candidate from contributing to its own reference, we construct a leave-one-manifold-out consensus: %
\begin{equation}
    b_i^{(-m,k,w)}
    {=} \argmax_{c\in[C]} \frac{1}{\lvert \gM \rvert - 1} \sum_{m'\in \gM \setminus \{m\}} p_{i,c}^{(m',k,w)}.
    \label{eq:leave-one-out-consensus}
\end{equation}
Let $\gY_c =\{i\in[N]:\hat y_i^{(\mathrm{SP})}=c\}$ be the initial pseudo-class groups, and let $\gY_{\mathrm{obs}} =\{c\in[C]:|\gY_c|>0\}$. We score each candidate by its mean within-group agreement:
\begin{equation}
    \kern-0.8em\gamma_m^{(k,w)}
    {=}
    \frac{1}{|\gY_{\mathrm{obs}}|}
    \kern-0.1em\sum_{c\in\gY_{\mathrm{obs}}}
    \kern-0.2em\frac{1}{|\gY_c|}
    \kern-0.1em\sum_{i\in\gY_c}
    \kern-0.2em\mathbbm{1}\!\bigl[
        \hat y_i^{(m,k,w)}
        {=}
        b_i^{(-m,k,w)}
    \bigr]\kern-0.2em.
    \label{eq:manifold-agreement}
\end{equation}
Averaging over fixed sets $\mathcal{K}{=}\{10,20,40,80\}$ and $\mathcal{W}{=}\{\mathrm{uniform}, \mathrm{confidence\text{-}distance}\}$ yields
\begin{equation}
    \gamma_m
    {=}
    \frac1{|\mathcal{K}||\mathcal{W}|}
    \sum_{k\in\mathcal{K}}\sum_{w\in\mathcal{W}}
    \gamma_m^{(k,w)},
    \quad
    \widehat m\kern-0.1em\in\kern-0.1em\arg\max_{m\in\mathcal{M}}\gamma_m,
    \label{eq:manifold-selection}
\end{equation}
the selected manifold $\widehat{m}$ (used in \cref{eq:denoised-logit}).\ %
This ground-truth-free criterion measures cross-manifold agreement rather than correctness, serving as a surrogate for the final denoising refinement quality. 

\subsection{Candidate Constrained Semantic Recovery}\label{sec:method-semantic-recovery}
While denoising consistently improves predictions under domain shift (\cref{tab:component-attribution-compact} and \cref{fig:promd-cross-benchmark-hyp-sensitivity-clip}), residual prediction errors remain. We try to recover these by querying an additional \gls{lvlm}. %
This step is optional. 

\subsubsection{Fragile Subset Selection}\label{sec:method-semantic-recovery-fragile-selection}
Without target annotations, residual prediction errors must be inferred from observable signals. Entropy and classification margins quantify uncertainty, but cannot reliably separate incorrect from correct but low-confidence predictions.\ We formulate an alternative selection strategy, which isolates potential residual errors by only considering samples where denoised predictions disagree with their nearest-neighbor proxy consensus.\ Let $\hat y_i^{(\mathrm{SP})}$ denote the initial prediction. For the selected manifold $\widehat m$, we define the denoised and nearest-neighbor predictions as 
$\hat y_i^{(\mathrm{D})}$ and $\hat y_i^{(\mathrm{M})}$, respectively.
We consider a sample \emph{fragile} when these two predictions disagree, giving $\gF {=} \bigl\{i\in[N]: \hat y_i^{(\mathrm{D})} \neq \hat y_i^{(\mathrm{M})} \bigr\}$.
As $\gF$ identifies disagreeing instance-neighborhood-evidence and \textbf{not} prediction quality, it may include correct predictions and exclude erroneous ones.\ %
Recovery may therefore introduce additional errors by modifying already correct predictions.\
To prioritize samples we propose a \emph{proxy-error-score} that combines label-independent signals and assigns higher values to samples whose zero-shot predictions change, exhibit high uncertainty or small decision margin on either the semantic proxy or the denoised logits.\ %
We retain the highest-scoring $(1\,{-}\,q)$ fraction of $\gF$, which forms $\gF_q\subseteq\gF$ $\left(q\in[0,1)\right)$. Larger $q$ retains fewer samples, reducing recovery cost, whereas smaller $q$ risks exposures of more correct predictions.

\begin{table*}[t]
  \centering
  \caption{Top-1 accuracy results on the \gls{sfda}-benchmark (in \%; macro accuracy for \gls{visdac}). $\Delta$ is measured against DIFO. $\dagger$ indicates missing or incomplete paper results; $\S$ indicates results from~\citet{tang_source-free_2024}. SOTA uses Dv2+Dv3; \gls{promd} uses selected $\widehat m$ per domain.}
  \label{tab:sfda-benchmark-results}
  \small
  \setlength{\tabcolsep}{3pt}
  \adjustbox{max width=\textwidth}{\begin{tabular}{llcccc cccccc cccccc ccc cc}
    \toprule
    \midrule
    \multirow{2}{*}{Method} & \multirow{2}{*}{Venue} & \multirow{2}{*}{$f_\theta$} & \multirow{2}{*}{VLM} & \multirow{2}{*}{Diff.} &
    \multirow{2}{*}{$f_S$} & \multicolumn{6}{c}{\glsxtrshort{domainnet}} &
    \multicolumn{6}{c}{\glsxtrshort{oh}} & \multicolumn{3}{c}{\glsxtrshort{visdac}} & \multicolumn{2}{c}{Global}\\
    \cmidrule(lr){7-10}\cmidrule(lr){11-12}\cmidrule(lr){13-16}\cmidrule(lr){17-18}\cmidrule(lr){19-21}\cmidrule(lr){22-23}
    & & & & &
    & \glsxtrshort{dnC} & \glsxtrshort{dnP} & \glsxtrshort{dnR} & \glsxtrshort{dnS} & Avg. & $\Delta$
    & \glsxtrshort{ohar} & \glsxtrshort{ohcl} & \glsxtrshort{ohpr} & \glsxtrshort{ohrw} & Avg. & $\Delta$
    & Top-1 & Macro & $\Delta$
    & Avg. $(\uparrow)$ & Rank $(\downarrow)$\\
    \midrule
    Source-Only$^{\diamond}$ & -- & -- & $\times$ & $\times$ & \checkmark
    & $53.87$ & $52.01$ & $64.55$ & $46.34$ & $54.19$ & $-25.89$
    & $56.13$ & $42.87$ & $68.17$ & $70.25$ & $59.36$ & $-23.72$
    & $53.44$ & $47.66$ & $-42.67$
    & $53.74$ & $16$\\
    \cmidrule{2-22}
    SHOT~\cite{liang_we_2020} & ICML20 & -- & $\times$ & $\times$ & \checkmark
    & $68.60$ & $65.00$ & $79.20$ & $59.70$ & $68.13$ & $-11.95$
    & $69.57$ & $56.93$ & $80.20$ & $80.60$ & $71.83$ & $-11.25$
    & -- & $82.70$ & $-7.63$
    & $74.22$ & $15$\\
    NRC~\cite{yang_exploiting_2021} & NeurIPS21 & -- & $\times$ & $\times$ & \checkmark
    & $65.67^{\S}$ & $65.93^{\S}$ & $79.03^{\S}$ & $59.23^{\S}$ & $67.47^{\S}$ & $-12.61$
    & $68.13$ & $57.57$ & $81.90$ & $81.20$ & $72.20$ & $-10.88$
    & -- & $85.90$ & $-4.43$
    & $75.19$ & $14$\\
    CoWA~\cite{lee_confidence_2022} & ICML22 & -- & $\times$ & $\times$ & \checkmark
    & $68.07^{\S}$ & $65.87^{\S}$ & $80.10^{\S}$ & $60.40^{\S}$ & $68.61^{\S}$ & $-11.47$
    & $69.87$ & $58.20$ & $80.97$ & $81.10$ & $72.53$ & $-10.55$
    & -- & $86.90$ & $-3.43$
    & $76.01$ & $13$\\
    RGV~\cite{zhu_revisiting_2025} & CVPR25 & -- & $\times$ & $\times$ & \checkmark
    & $74.75^\dagger$ & $72.05^\dagger$ & $83.20^\dagger$ & $67.65^\dagger$ & $74.41^\dagger$ & $-5.67$
    & $70.63$ & $60.80$ & $82.60$ & $82.50$ & $74.13$ & $-8.95$
    & -- & $89.10$ & $-1.23$
    & $79.21$ & $12$\\
    \cmidrule{3-22}
    DPTM~\cite{huang_diffusion-driven_2025} & NeurIPS25 & -- & $\times$ & \checkmark & \checkmark
    & $85.47$ & $85.47$ & $90.83$ & $79.00$ & $85.19$ & $+5.11$
    & $\underline{90.87}$ & $\underline{86.73}$ & $\underline{94.30}$ & $92.73$ & $\underline{91.16}$ & $+8.08$
    & -- & \textcolor{cvprblue}{$\mathbf{97.60}$} & $+7.27$
    & $\underline{91.32}$ & $\underline{2}$\\
    \cmidrule{3-22}
    CLIP$^{\diamond}$~\cite{radford_learning_2021} & ICML21
    &
    & \checkmark & $\times$ & $\times$
    & $79.12$ & $77.60$ & $90.17$ & $74.65$ & $80.38$ & $+0.30$
    & $78.70$ & $65.43$ & $86.51$ & $87.88$ & $79.63$ & $-3.45$
    & $85.50$ & $88.05$ & $-2.28$
    & $82.69$ & $11$\\
    DIFO~\cite{tang_source-free_2024} & CVPR24 & & \checkmark & $\times$ & \checkmark
    & $80.65$ & $77.06$ & $87.04$ & $75.56$ & $80.08$ & $\pm 0.00$
    & $82.27$ & $70.40$ & $90.80$ & $88.83$ & $83.08$ & $\pm 0.00$
    & -- & $90.33$ & $\pm 0.00$
    & $84.50$ & $9$\\
    ProDe~\cite{tang_proxy_2025} & ICLR25 & & \checkmark & $\times$ & \checkmark
    & $85.33$ & $83.23$ & $92.37$ & $79.13$ & $85.02$ & $+4.94$
    & $82.67$ & $72.60$ & $92.00$ & $90.67$ & $84.48$ & $+1.40$
    & -- & $91.00$ & $+0.67$
    & $86.83$ & $6$\\
    DUET~\cite{lee_duet_2025} & NeurIPS25 & & \checkmark & $\times$ & \checkmark
    & $82.73$ & $80.33$ & $89.67$ & $79.00$ & $82.93$ & $+2.85$
    & $83.30$ & $73.77$ & $90.77$ & $91.03$ & $84.72$ & $+1.64$
    & -- & $91.40$ & $+1.07$
    & $86.35$ & $7$\\
    VSFOT~\cite{han_vision-language_2026} & CVPR26 & & \checkmark & $\times$ & \checkmark
    & $83.26$ & $79.63$ & $90.44$ & $77.20$ & $82.63$ & $+2.55$
    & $83.39$ & $74.76$ & $91.80$ & $91.40$ & $85.34$ & $+2.26$
    & -- & $90.57$ & $+0.24$
    & $86.18$ & $8$\\
    ProS$^2$T~\cite{zhou_pros2t_2026} & ArXiv26 & & $\checkmark$ & $\times$ & $\times$
    & $86.00$ & $83.60$ & $92.80$ & $80.00$ & $85.60$ & $+5.52$
    & $83.50$ & $74.40$ & $92.80$ & $90.80$ & $85.40$ & $+2.32$
    & -- & $91.10$ & $+0.77$
    & $87.37$ & $5$\\
    \cmidrule{4-22}
    ECALP$^{\diamond}$~\cite{li_efficient_2025} & ICLR25 & & \checkmark & $\times$ & $\times$
    & $79.38$ & $78.14$ & $90.90$ & $75.13$ & $80.89$ & $+0.81$
    & $79.27$ & $68.61$ & $88.06$ & $88.57$ & $81.13$ & $-1.95$
    & $85.92$ & $88.70$ & $-1.63$
    & $83.57$ & $10$\\
    SOTA$^{\diamond}$~\cite{hu_sota_2026} & CVPR26 & & \checkmark & $\times$ & $\times$
    & $87.91$ & $85.34$ & $93.87$ & $84.34$ & $87.86$ & $+7.78$
    & $90.28$ & $85.43$ & $93.92$ & $\underline{93.89}$ & $90.88$ & $+7.80$
    & $88.04$ & $90.56$ & $+0.23$
    & $89.77$ & $4$\\
    \rowcolor{gray!17} \gls{promd}$^{\diamond}$ & -- & & \checkmark & $\times$ & $\times$
    & $\underline{90.40}$ & $\underline{88.00}$ & $\underline{94.28}$ & $\mathbf{87.79}$ & $\underline{90.12}$ & $+10.04$
    & $88.50$ & $80.99$ & $94.23$ & $93.80$ & $89.38$ & $+6.30$
    & $\underline{90.48}$ & $92.38$ & $+2.05$
    & $90.63$ & $3$\\
    \rowcolor{gray!17} \gls{promd}+$^{\diamond}$ & -- & \multirow{-10}{*}{\begin{turn}{90}$\cdots\,\,\,\,$ViT-B/32$\quad \cdots$\end{turn}} & \checkmark & $\times$ & $\times$
    & $\mathbf{91.50}$ & $\mathbf{89.06}$ & $\mathbf{95.75}$ & $\underline{87.60}$ & $\mathbf{90.98}$ & $+10.90$
    & $\mathbf{91.92}$ & $\mathbf{90.56}$ & $\mathbf{95.61}$ & $\mathbf{95.34}$ & $\mathbf{93.36}$ & $+10.28$
    & \textcolor{cvprblue}{$\mathbf{90.63}$} & $\underline{92.58}$ & $+2.25$
    & $\mathbf{92.31}$ & $\mathbf{1}$\\
    \cmidrule{2-23}
    \gls{siglip2}$^{\diamond}$~\cite{tschannen_siglip_2025} &
    &
    & \checkmark & $\times$ & $\times$
    & $93.80$ & $93.03$ & $96.57$ & $92.21$ & $93.90$ & $+13.82$
    & $94.27$ & $92.05$ & $98.60$ & \textcolor{cvprblue}{$\mathbf{97.43}$} & $95.59$ & $+12.51$
    & $86.20$ & $89.90$ & $-0.43$
    & $93.13$ & $3$\\
    ECALP$^{\diamond}$~\cite{li_efficient_2025} & ICLR25 & & $\checkmark$ & $\times$ & $\times$
    & $90.69$ & $88.34$ & $95.51$ & $88.35$ & $90.72$ & $+10.64$
    & $91.80$ & $92.16$ & $98.09$ & $96.56$ & $94.65$ & $+11.57$
    & $85.07$ & $88.99$ & $-1.34$
    & $91.45$ & $5$\\
    SOTA$^{\diamond}$~\cite{hu_sota_2026} & CVPR26 & & \checkmark & $\times$ & $\times$
    & $92.90$ & $92.26$ & $96.44$ & $90.89$ & $93.12$ & $+13.04$
    & $94.97$ & $92.44$ & $98.00$ & $96.56$ & $95.49$ & $+12.41$
    & $87.00$ & $90.29$ & $-0.04$
    & $92.97$ & $4$\\
    \rowcolor{gray!17}\gls{promd}$^{\diamond}$ & -- &
    & \checkmark & $\times$ & $\times$
    & \textcolor{cvprblue}{$\mathbf{93.82}$}
    & \textcolor{cvprblue}{$\mathbf{93.45}$}
    & $96.54$
    & \textcolor{cvprblue}{$\mathbf{92.74}$}
    & \textcolor{cvprblue}{$\mathbf{94.14}$}
    & $+14.06$
    & $93.61$ & $92.74$
    & \textcolor{cvprblue}{$\mathbf{98.72}$}
    & $96.86$ & $95.48$ & $+12.40$
    & $89.44$ & $92.09$ & $+1.76$
    & $93.90$ & $2$\\
    \rowcolor{gray!17}\gls{promd}+$^{\diamond}$ & -- & \multirow{-5}{*}{\begin{turn}{90}\gls{siglip2}\end{turn}}
    & \checkmark & $\times$ & $\times$
    & $93.69$ & $92.81$
    & \textcolor{cvprblue}{$\mathbf{96.80}$}
    & $91.61$ & $93.73$ & $+13.65$
    & \textcolor{cvprblue}{$\mathbf{95.06}$}
    & \textcolor{cvprblue}{$\mathbf{93.40}$}
    & $98.58$ & $97.36$
    & \textcolor{cvprblue}{$\mathbf{96.10}$}
    & $+13.02$
    & $89.23$ & $91.94$ & $+1.61$
    & \textcolor{cvprblue}{$\mathbf{93.92}$} & $\mathbf{1}$\\
    \midrule
    \bottomrule
  \end{tabular}}
\end{table*}

\subsubsection{Semantic Recovery}
\label{sec:semantic-recovery-stage}
Following candidate-restricted recognition in RAR~\cite{liu_rar_2024} and ProtoRAG~\cite{wang_protorag_2026}, we constrain semantic recovery to a reduced label set.\ We facilitate semantic recovery on $\gF_q$ by restricting the label space for each selected sample to the top-$k_\gC$ \gls{promd} candidates:
\begin{equation}
    \mathcal{C}_i
    =
    \operatorname*{arg\,topK}_{c\in[C]}
    \widetilde{\ell}_{i,c}^{(\widehat m, k)},
    \qquad i\in\gF_q.
\label{eq:candidate-set}
\end{equation}
We apply \emph{candidate-forced} prompting, which allows a generative \gls{lvlm} to be used as a discriminative scorer rather than autoregressively generating unrestricted class names. This ensures that recovery stays within the denoised candidate set. For each sample $i\in\gF_q$, we pass $x_i$ together with the candidate labels and their canonical identifiers (e.g., \texttt{C00}). Let $a_c{=} (a_{c,l})^{L_c}_{l=1}$ %
encode the identifier of candidate $c\in\mathcal{C}_i$, and let $t_i^{(\mathrm{Rec})}$ be the candidate-constrained prompt, then the length-normalized score is
\begin{equation}
    s_{i,c}^{(\mathrm{Rec})}
    {=}
    \frac{1}{L_c}
    \sum_{l=1}^{L_c}%
    \log p^{(\mathrm{Qwen})}
    \bigl(
        a_{c,l}
        \mid
        x_i,t_i^{(\mathrm{Rec})},a_{c,<l}
    \bigr)\kern-0.1em.
\label{eq:teacher-forced-score}
\end{equation}
We normalize these scores within $\mathcal{C}_i$ to get $\hat y_i^{(\mathrm{Rec})}$. %
\gls{promd} final predictions use $\hat y_i^{(\mathrm{Rec})}$ for $i \in \gF_{q}$ and $\hat y_i^{(\mathrm{D})}$ otherwise.

\section{Experiments}\label{sec:experiments}

We compare \gls{promd} against transductive \gls{vlm} refinement methods and against conventional, diffusion-based, and \gls{vlm}-guided \gls{sfda} approaches. %
\gls{promd} is evaluated on a \gls{sfda} benchmark (SFDABench) and a categorization benchmark (CatBench), which comprise 20 domains/tasks (14 datasets). We emphasize comparisons with ECALP~\cite{li_efficient_2025} and SOTA~\cite{hu_sota_2026}, the closest related baselines.\ For SOTA, \cref{tab:categorization-benchmark-results} reports its best published configuration (\gls{dinov2}+\gls{dinov3}). We apply ECALP's and SOTA's published CatBench configuration unchanged to SFDABench.\ Primary comparisons use ViT-B \gls{clip} backbones; additional results use \gls{resnet50}
and \gls{siglip2}~\cite{google_siglip-2_2026,ilharco_openclip_2021}.

We report published results where available; ${\diamond}$ marks results obtained with our codebase on a single NVIDIA RTX PRO 6000 Blackwell.\ \gls{promd} uses prediction-backbone-specific denoising profiles.\ We share one denoising profile per backbone across CatBench. For SFDABench, we share one profile across domains per dataset, where selection is guided by dataset size and expected backbone quality. \gls{promd}${+}$ employs the same profiles.\ We fix recovery quantile at $q{=}0.5$.\ SFDABench uses generic prompts with mean-prompt aggregation (V5),
whereas CatBench follows DMN's~\cite{zhang_dual_2024} \gls{cupl}-style prompting~\cite{pratt_what_2023}.

\begin{table*}[h]
    \caption{Top-1 accuracy results on the categorization benchmark (in \%). $\dagger$ reports from \citet{li_efficient_2025}, $\sharp$ exhaustively searches the fusion hyperparameter with testing-data ground truth on each individual downstream task, $^{**}$ uses additional unlabeled training data, $\ddagger$ indicates zero-shot but not training-free. $\Delta$ is measured against ECALP. SOTA uses Dv2+Dv3; \gls{promd} uses selected manifold $\widehat m$ per dataset. %
    }
    \label{tab:categorization-benchmark-results}
    \centering
    \adjustbox{max width=\textwidth}{
    \begin{tabular}{llc|*{11}{c}|ccc}
         \toprule
         \midrule
         Method & Venue & $f_\theta$ &
         \glsxtrshort{aircraft} & \glsxtrshort{caltech} & \glsxtrshort{car} & \glsxtrshort{dtd} & \glsxtrshort{eurosat} & \glsxtrshort{flowers} & \glsxtrshort{food} & \glsxtrshort{imagenet} & \glsxtrshort{pets} & \glsxtrshort{sun} & \glsxtrshort{ucf} & Avg. $(\uparrow)$ & $\Delta (\uparrow)$ & Rank $(\downarrow)$ \\
         \midrule
         CLIP$^{\diamond}$~\cite{radford_learning_2021} & ICML21
         &
         & $27.21$ & $94.12$ & $66.16$ & $53.25$ & $55.57$ & $73.04$ & $86.09$ & $69.93$ & $90.32$ & $67.98$ & $70.47$
         & $68.56$ & $-1.98$ & $9$ \\
         DMN$^\dagger$~\cite{zhang_dual_2024} & CVPR24 & &
         $28.29$ & $93.63$ & $67.01$ & $54.85$ & $56.22$ & $75.32$ & $84.05$ & $70.51$ & $91.22$ & $69.14$ & $71.95$
         & $69.29$ & $-1.25$ & $8$ \\
         DMN$^{\dagger, \sharp}$~\cite{zhang_dual_2024} & CVPR24 & &
         $30.03$ & $95.38$ & $67.96$ & $55.85$ & $59.43$ & $74.49$ & $85.08$ & $72.25$ & $92.04$ & $70.18$ & $72.51$
         & $70.47$ & $-0.07$ & $6$ \\
         ZLaP$^{\dagger, **}$~\cite{stojnic_label_2024} & CVPR24 & &
         $25.44$ & $93.06$ & $65.63$ & $48.58$ & $55.62$ & $73.49$ & $86.92$ & $70.17$ & $87.14$ & $67.44$ & $71.45$
         & $67.72$ & $-2.82$ & $10$ \\
         TransCLIP-ZS~\cite{zanella_boosting_2024} & NeurIPS24 & &
         $26.90$ & $92.70$ & $69.40$ & $49.50$ & $65.10$ & $76.70$ & $87.10$ & $70.30$ & $92.60$ & $68.90$ & $74.40$
         & $70.33$ & $-0.21$ & $7$ \\
         ECALP$^{\dagger}$~\cite{li_efficient_2025} & ICLR25 & &
         $29.49$ & $94.40$ & $68.20$ & $56.32$ & $56.53$ & $75.96$ & $85.72$ & $71.26$ & $92.31$ & $70.35$ & $75.44$
         & $70.54$ & $\pm 0.00$ & $5$ \\
         GTA-CLIP$^{\ddagger}$~\cite{saha_generate_2025} & ICCV25 & &
         $29.30$ & $95.50$ & $72.10$ & $58.50$ & $76.40$ & $82.10$ & $87.40$ & $71.90$ & $93.40$ & $\underline{73.50}$ & $79.10$
         & $74.50$ & $+3.96$ & $4$ \\
         SOTA~\cite{hu_sota_2026} & CVPR26 & &
         $34.00$ & $\underline{96.60}$ & $\underline{78.50}$ & $55.90$
         & \textcolor{cvprblue}{$\mathbf{87.20}$}
         & $\underline{84.50}$ & $90.00$ & $\underline{79.90}$ & $\underline{95.30}$ & $72.80$ & $\underline{81.10}$
         & $\underline{77.80}$ & $+7.26$ & $\underline{2}$ \\
         \rowcolor{gray!17} \gls{promd}$^{\diamond}$  & -- & &
         $\underline{37.26}$ & $95.86$ & $77.22$ & $\underline{58.81}$ & $64.73$ & $80.76$ & $\underline{93.77}$ & $78.02$ & $93.00$ & $71.64$ & $76.53$
         & $75.23$ & $+4.69$ & $3$ \\
         \rowcolor{gray!17} \gls{promd}+$^{\diamond}$  & -- & \multirow{-10}{*}{\begin{turn}{90}\small$\cdots\quad\,$ViT-B/16$\quad\cdots$\end{turn}} &
         $\mathbf{48.09}$ & $\mathbf{98.13}$ & $\mathbf{83.27}$ & $\mathbf{63.89}$ & $\underline{83.68}$ & $\mathbf{87.90}$ & $\mathbf{93.93}$ & $\mathbf{82.70}$ & $\mathbf{96.35}$ & $\mathbf{75.96}$ & $\mathbf{83.03}$
         & $\mathbf{81.54}$ & $+11.00$ & $\mathbf{1}$ \\
         \cmidrule(l){1-17}
         \gls{siglip2}$^{\diamond}$~\cite{tschannen_siglip_2025} &
         &
         & $68.95$ & $98.66$ & $95.54$ & $72.22$ & $50.16$ & $87.74$ & $94.89$ & $84.24$ & $96.40$ & $77.30$ & $85.49$
         & $82.87$ & $-0.65$ & $5$ \\
         ECALP$^{\diamond}$~\cite{li_efficient_2025} & ICLR25 & &
         \textcolor{cvprblue}{$\mathbf{71.50}$}
         & $98.05$ & $95.44$ & $71.34$ & $54.80$ & $89.85$ & $95.03$ & $82.48$ & $96.10$ & $77.01$ & $87.18$
         & $83.52$ & $\pm 0.00$ & $4$ \\
         SOTA$^{\diamond}$~\cite{hu_sota_2026} & CVPR26 & &
         
        $63.22$ & $98.82$ & $95.00$ & $70.92$ & $61.46$ & $95.17$ & $93.74$ & $84.79$ & $96.54$ & $78.81$ & \textcolor{cvprblue}{$\mathbf{89.27}$} & $84.34$ & $+0.82$ & $2$ \\
         \rowcolor{gray!17}\gls{promd}$^{\diamond}$ & -- &
         & $66.46$ & $98.54$ & $95.35$ & $74.47$ & $55.22$ & $88.71$ & $95.04$ & $84.21$ & $96.27$ & $78.11$ & $86.68$
         & $83.55$ & $+0.03$ & $3$ \\
         \rowcolor{gray!17}\gls{promd}+ $^{\diamond}$ & -- & \multirow{-5}{*}{\begin{turn}{90} \small \gls{siglip2}\end{turn}}
         & $70.03$
         & \textcolor{cvprblue}{$\mathbf{98.90}$}
         & \textcolor{cvprblue}{$\mathbf{95.80}$}
         & \textcolor{cvprblue}{$\mathbf{77.07}$}
         & $72.70$
         & \textcolor{cvprblue}{$\mathbf{98.01}$}
         & \textcolor{cvprblue}{$\mathbf{95.23}$}
         & \textcolor{cvprblue}{$\mathbf{86.07}$}
         & \textcolor{cvprblue}{$\mathbf{96.73}$}
         & \textcolor{cvprblue}{$\mathbf{81.10}$}
         & $88.61$
         & \textcolor{cvprblue}{$\mathbf{87.30}$}
         & $+3.78$ & \textcolor{cvprblue}{$\mathbf{1}$}\\
         \midrule
         \bottomrule
    \end{tabular}
    }
\end{table*}

\subsection{\glsxtrshort{sfda} and Categorization Benchmarks}\label{sec:experiments-sfda-categorization}
\cref{tab:sfda-benchmark-results} compares \gls{promd}'s predictions directly with \emph{\textbf{existing \gls{sfda} adaptation methods}}. VLM, Diff., and $f_S$ indicate use of a \gls{vlm}, diffusion, and a task-specific source-trained model, respectively.\ With ViT-B/32, \gls{promd} exceeds SOTA~\cite{hu_sota_2026} by $+\mathbf{0.86}$ \gls{pp} in average accuracy, while trailing DPTM. \gls{promd}${+}$ further improves SOTA by $+\mathbf{2.54}$ \gls{pp}, achieving the highest average accuracy, exceeding DPTM by $+\mathbf{0.99}$ \gls{pp}. DPTM retains the best accuracy on \gls{visdac}. With \gls{siglip2}, \gls{promd}${+}$ reaches $93.92\%$, only $+0.02$ \gls{pp} above \gls{promd}. 
\cref{tab:categorization-benchmark-results} compares \emph{\textbf{transductive \gls{vlm} refinement methods}} across 11 datasets. With \gls{clip}, \gls{promd} improves zero-shot average accuracy by $\mathbf{+6.67}$ \gls{pp}, exceeding GTA-CLIP by $\mathbf{+0.73}$ \gls{pp} and ECALP by $\mathbf{+4.69}$ \gls{pp}. SOTA outperforms \gls{promd} by $+2.57$ \gls{pp}. \gls{promd}${+}$ achieves the highest accuracy, improving SOTA by $\mathbf{+3.74}$ \gls{pp} and ranking first on 10/11 datasets within the \gls{clip} backbone. For \gls{siglip2}, \gls{promd} and ECALP perform on par, improving the zero-shot baseline by $\sim\mathbf{0.65}$ \gls{pp}, while \gls{promd}${+}$ exceeds SOTA by $\mathbf{+2.96}$ \gls{pp} and zero-shot by $\mathbf{+4.43}$ \gls{pp}.\ 

These results show that the benefits of \gls{promd} extend beyond zero-shot predictions, further improving accuracy even after prior transductive refinement.

\subsection{Downstream Classifier Training}\label{sec:experiments-target-classifier-training}
We evaluate whether \gls{promd}'s predictions provide effective supervision for separate classifier training, realizing the distillation setting proposed under \gls{sida}. We train \gls{resnet} classifiers with \gls{ce} on \gls{promd}'s final hard pseudo-labels and optionally add \gls{kl} distillation from the denoised class distribution. \cref{tab:sida-results-main-clip} compares zero-shot \gls{clip} and \gls{promd} supervision, alongside SHOT- and DIFO-derived variants adapted to operate within \gls{sida} (denoted by $\dagger$). \cref{tab:sida-results-main-clip} demonstrates: zero-shot \gls{clip} supervision achieves decent performance; \gls{kl} distillation improves accuracy over \gls{ce} alone; \gls{vlm} refinement transfers to target-classifier learning (cf. \gls{promd} vs. \gls{clip}); \gls{promd} supervision improves accuracy by $+5.44$\gls{pp} over zero-shot under the \gls{ce}+\gls{kl} objective; and
\gls{promd}'s distilled classifiers outperform the \gls{vlm}-guided \gls{sfda} method. %
Evaluated SHOT and DIFO \gls{sida}-adjusted variants yield no gains over simple \gls{ce}+\gls{kl} distillation.

\begin{table}[t]
    \centering
    \caption{Classifier training (in \gls{sida}) reporting the same \gls{resnet} backbones as \gls{sfda} methods in \cref{tab:sfda-benchmark-results}, trained for 15 epochs. Best validation accuracy in parentheses. }
    \label{tab:sida-results-main-clip}
    \adjustbox{max width=\linewidth}{\begin{tabular}{lccccc}
        \toprule
        \midrule
        Method & \glslink{domainnet}{DN} & \glslink{oh}{OH} & \glslink{visdac}{V} & Avg. & Rank \\
        \midrule
        Source-Only$^\diamond$ & $54.19$ & $59.36$ & $47.66$ & $53.74$ & $11\;(11)$ \\
        SHOT~\cite{liang_we_2020} & $68.13$ & $71.83$ & $82.70$ & $74.22$ & $9\;(10)$ \\
        DIFO~\cite{tang_source-free_2024} & $80.08$ & $83.08$ & $90.33$ & $84.50$ & $6\;(8)$ \\
        ProDe~\cite{tang_proxy_2025} & $85.02$ & $84.48$ & $91.00$ & $86.83$ & $5\;(5)$ \\
        ProS$^2$T~\cite{zhou_pros2t_2026} & $85.60$ & $85.40$ & $\underline{91.10}$ & $87.37$ & $4\;(4)$ \\
        \cmidrule{2-6}
        \gls{clip} (CE)$^\diamond$ & $80.83\;(80.88)$ & $81.13\;(81.33)$ & $89.03\;(89.11)$ & $83.66\;(83.77)$ & $8\;(9)$ \\
        \gls{clip} (CE+KL)$^\diamond$ & $81.86\;(81.99)$ & $81.43\;(81.88)$ & $89.84\;(89.90)$ & $84.38\;(84.59)$ & $7\;(7)$ \\
        \gls{promd} (CE)$^\diamond$ & $\underline{87.10}\;(87.16)$ & $87.40\;(87.66)$ & $91.01\;(91.01)$ & $\underline{88.50}\;(88.61)$ & $\underline{2}\;(\underline{2})$ \\
        \gls{promd} (CE+KL)$^\diamond$ & $\textbf{89.02}\;(89.13)$ & $\textbf{88.62}\;(88.69)$ & $\textbf{91.82}\;(91.82)$ & $\textbf{89.82}\;(89.88)$ & $\mathbf{1}\;(\mathbf{1})$ \\
        \cmidrule{2-6}
        SHOT-IM + (CE)$^{\dagger,\diamond}$ & $87.06\;(87.14)$ & $87.39\;(87.63)$ & $91.06\;(91.06)$ & $\underline{88.50}\;(88.61)$ & $2\;(2)$ \\
        DIFO + (CE)$^{\dagger,\diamond}$ & $68.87\;(80.96)$ & $\underline{87.97}\;(88.04)$ & $40.66\;(88.23)$ & $65.83\;(85.74)$ & $10\;(6)$ \\
        \cmidrule{2-6}
        \gls{clip}$^\diamond$ & $80.38$ & $79.63$ & $88.05$ & $82.69$ & -- \\
        \gls{promd}$^\diamond$ & $90.12$ & $89.38$ & $92.38$ & $90.63$ & -- \\
        CE with GT$^{\diamond}$ & $92.23\;(92.25)$ & $93.87\;(93.98)$ & $94.27\;(94.27)$ & $93.46\;(93.50)$ & -- \\
        \midrule
        \bottomrule
    \end{tabular}}
\end{table}

\subsection{Better Predictions Limit Refinement Headroom}\label{sec:experiments-discussion}
Refinement provides smaller improvements when the initial zero-shot predictions are already strong (cf. \cref{tab:sfda-benchmark-results,tab:categorization-benchmark-results}). %
Consider $\boldsymbol\ell_i^{(t)}=\rvu_i+t\boldsymbol\xi_i$, where $t\ge0$ controls prediction-noise magnitude. We define
\begin{align}
    R_i^{(t)}(\lambda_{\mathrm{den}})
        &:=\mathbb E\|
        \widetilde{\boldsymbol\ell}_i^{(m,k,t)}-\rvu_i
        \|_2^2, \label{eq:expected-squarred-error-denoising-theory}\\
    G_i^{(t)}(\lambda_{\mathrm{den}})
        &:=R_i^{(t)}(0)-R_i^{(t)}(\lambda_{\mathrm{den}}).
        \label{eq:gain-denoising-theory}
\end{align}
where $R$ measures expected squared error of the denoised logit and $G$ its reduction through denoising. Here, $\widetilde{\boldsymbol\ell}_i^{(m,k,t)}$ denotes the logit denoised with $\lambda_{\mathrm{den}}$. %

\begin{table}[t]
  \centering
  \small
  \caption{Cumulative component attribution of \gls{promd} (in \%).%
  }
  \label{tab:component-attribution-compact}
  \adjustbox{max width=\linewidth}{\begin{tabular}{lcccc}
    \toprule
    \midrule
    \multirow{2}{*}{\gls{promd} stages} & \multicolumn{2}{c}{SFDABench (3 datasets)} & \multicolumn{2}{c}{CatBench (11 datasets)} \\
    \cmidrule(lr){2-3}\cmidrule(lr){4-5}
    & \gls{clip} & \gls{siglip2} & \gls{clip} & \gls{siglip2} \\
    \midrule
    Zero-shot %
    & $81.70$ & $92.76$ & $68.55$ & $82.87$ \\
    \hspace{0.5em}+ Prompt & $82.69$ ($+0.99$) & $93.13$ ($+0.37$) & --$^1$ & --$^1$ \\
    \hspace{0.5em}+ Denoising & $89.67$ ($\mathbf{+6.98}$) & $93.89$ ($\mathbf{+0.76}$) & $74.80$ ($\mathbf{+6.25}$) & $83.39$ ($+0.52$) \\
    \hspace{1.4em}\textcolor{gray}{Oracle config} & \textcolor{gray}{$90.03$ ($+0.36$)} & \textcolor{gray}{$93.95$ ($+0.06$)} & \textcolor{gray}{$75.13$ ($+0.33$)} & \textcolor{gray}{$84.24$ ($+0.85$)} \\
    \hspace{0.5em}+ Semantic recovery & $\mathbf{90.63}$ ($+0.95$) & $\mathbf{93.90}$ ($+0.01$) & $\mathbf{75.23}$ ($+0.44$) & $\mathbf{83.55}$ ($+0.16$) \\
    \hspace{1.4em}\textcolor{gray}{Oracle fragile set} & \textcolor{gray}{$94.76$ ($+4.13$)} & \textcolor{gray}{$96.35$ ($+2.45$)} & \textcolor{gray}{$82.94$ ($+7.71$)} & \textcolor{gray}{$88.09$ ($+4.54$)} \\
    \midrule
    \bottomrule
  \end{tabular}
  }
  \tiny $^1$ CatBench follows DMN's~\cite{zhang_dual_2024} CuPL strategy, thus prompting cannot contribute to any performance gain.
\end{table}
\begin{corollary}[Diminishing refinement gain]
\label{cor:main-diminishing-gains} Consider the noisy prediction model with $\widetilde{\boldsymbol\ell}_i^{(m,k,t)}$ and fix $\rvu_i$, $\lambda_{\mathrm{den}}$, nearest neighbors, and joint distribution of zero-mean noise with finite second moments. If $G_i^{(t_0)}(\lambda)>0$ for some noise level $t_0>0$, then $G_i^{(t_2)}(\lambda)<G_i^{(t_1)}(\lambda)$ whenever $0\le t_2<t_1$.
\end{corollary}

Hence, `cleaner' predictions ($\boldsymbol{\ell}_i {\to} \rvu_i$) have less noise to correct (neighbor mismatch persists), limiting achievable gain.\ Consequently, let $A_0(t)$ be the zero-shot and $A_{\lambda}(t)$ the denoised accuracy, then the accuracy gain is bounded by the fraction of initially incorrect predictions $1{-}A_0(t)$ with
\begin{equation}
    A_\lambda(t)-A_0(t)
        \le\sum\nolimits_i w_i\min\bigl\{
            1, 2t^2V_i / \bigl(\Delta_i^{(u)}\bigr)^2
        \bigr\}.
    \label{eq:class-decisions-accuracy-headroom-main-short}
\end{equation}
where $w_i\ge0$, $\sum_i w_i {=} 1$, and $V_i:=\mathbb{E}[\|\boldsymbol\xi_i\|_2^2]$ is the expected squared noise magnitude. The ideal-logit margin is $\Delta_i^{(u)}:=u_{i,y_i}-\max_{c\ne y_i}u_{i,c}>0$, assuming that $\rvu_i$ uniquely identifies the ground-truth class $y_i$. As $t{\to}0$, the zero-shot predictions become increasingly accurate as $1{-}A_0(t)$ reduces, leaving less room for denoising to improve performance.

\section{Ablations}\label{sec:ablations}
We present key ablations.

\subsection{\gls{promd} Component Performance Attribution}\label{sec:ablations-performance-attribution}
Starting from zero-shot predictions, \cref{tab:component-attribution-compact} shows that the employed prompt
improves SFDABench accuracy by ${+}0.99/{+}0.37$ \gls{pp} for \gls{clip}/\gls{siglip2}; CatBench follows DMN's~\cite{zhang_dual_2024} prompt protocol. Denoising accounts for the greatest improvements with ${+}6.98/{+}0.76$ \gls{pp}, with smaller gains for stronger predictors (cf. \cref{sec:experiments-discussion}). %
Recovery adds ${+}0.95/{+}0.01$ \gls{pp}, while the oracle fragile set offers a further ${+}4.13/{+}2.45$ \gls{pp}. The results on CatBench follow a similar pattern, with smaller denoising gains but larger oracle-recovery gains (${+}7.71/{+}4.54$ \gls{pp}).\ This identifies a bottleneck: label-free signals imperfectly distinguish incorrect predictions during residual-error selection, limiting recovery.\ The oracle quantities demonstrate post-hoc performance headroom. %

\subsection{Proxy-Manifold Selection}\label{sec:ablations-proxy-manifold-selection}
We evaluate our manifold selection strategy against an oracle that selects the highest-accuracy manifold at fixed $k$ and $\lambda_{\mathrm{den}}$. Across 20 domains/tasks (14 datasets), our proxy-manifold selector matches the oracle choice in 18 cases for \gls{clip} and 15 for \gls{siglip2}, with mean accuracy gaps of only $0.01$ and $0.04$ \gls{pp}, respectively. Thus, cross-manifold agreement provides an effective surrogate for manifold selection without target labels.\ We additionally evaluate against fixed-manifold configurations and find that the improvements over the best fixed manifold are modest. This suggests that manifold quality is more important than manifold selection itself, while demonstrating that no single representation is uniformly optimal across all datasets.

\subsection{Sensitivity of Denoising Hyperparameters}\label{sec:ablations-k-and-lambda-neighborhood}
\begin{figure}[t]
  \centering
  \pgfplotsset{
  colormap={gainmap}{
    rgb255(0cm)=(166,73,30);
    rgb255(1cm)=(215,151,108);
    rgb255(2cm)=(247,246,242);
    rgb255(3cm)=(116,181,175);
    rgb255(4cm)=(0,91,88)
  },
  gainaxis/.style={
    scale only axis,width=3.50cm,height=3.50cm,
    xmin=-0.5,xmax=3.5,ymin=-0.5,ymax=3.5,
    enlargelimits=false,y dir=reverse,
    axis on top,axis line style={black!30,line width=0.35pt},
    tick style={draw=none},
    xtick={0,1,2,3},xticklabels={3,5,10,20},
    ytick={0,1,2,3},yticklabels={0.1,0.25,0.5,0.75},
    tick label style={font=\fontsize{8}{10}\selectfont,text=black!80},
    xlabel={$k$},xlabel style={font=\fontsize{9}{11}\selectfont},
    title style={font=\bfseries\fontsize{10}{12}\selectfont,yshift=2pt},
    point meta min=-6.25,point meta max=6.25,colormap name=gainmap,
  },
}
\def\DrawGainTable#1#2#3{%
  \pgfplotstableread[col sep=space,row sep=\\]{x y gain\\#1}\gaindata
  \addplot[matrix plot*,mesh/cols=4,point meta=explicit,
    draw=white,line width=0.65pt]
    table[x=x,y=y,meta=gain]\gaindata;
  \pgfplotstableforeachcolumnelement{gain}\of\gaindata\as\gainvalue{%
    \pgfmathtruncatemacro{\cellx}{mod(\pgfplotstablerow,4)}
    \pgfmathtruncatemacro{\celly}{floor(\pgfplotstablerow/4)}
    \pgfmathtruncatemacro{\darkcell}{abs(\gainvalue)>=4.0}
    \ifnum\darkcell=1 \def\cellcolor{white}\else\def\cellcolor{black!85}\fi
    \def\cellweight{\mdseries}\def\cellmathweight{\unboldmath}
    \ifnum\cellx=#2 \ifnum\celly=#3 \def\cellweight{\bfseries}\def\cellmathweight{\boldmath}\fi\fi
    \edef\DrawGainCell{\noexpand\node[text=\cellcolor,
      font={\expandafter\noexpand\cellweight\expandafter\noexpand\cellmathweight\noexpand\fontsize{8}{10}\noexpand\selectfont},inner sep=0pt]
      at (axis cs:\cellx,\celly)
      {\noexpand\pgfmathprintnumber[fixed,fixed zerofill,precision=2]{\gainvalue}};}
    \DrawGainCell
  }
  \draw[white,line width=1.7pt]
    (axis cs:{#2-0.44},{#3-0.44}) rectangle (axis cs:{#2+0.44},{#3+0.44});
  \draw[black!80,line width=0.6pt]
    (axis cs:{#2-0.44},{#3-0.44}) rectangle (axis cs:{#2+0.44},{#3+0.44});
}
\adjustbox{max width=\linewidth}{%
\begin{tikzpicture}
\begin{groupplot}[
  gainaxis,group style={group size=4 by 1,horizontal sep=0.35cm},
]
\nextgroupplot[title={\gls{clip}},ylabel={$\lambda_\mathrm{den}$},ylabel style={font=\fontsize{9}{11}\selectfont, at={(axis description cs:0.015,0.5)},anchor=north}, title style={yshift=-5pt}, xlabel style={yshift=+5pt},]
\DrawGainTable{
  0 0 0.405752\\
  1 0 0.406375\\
  2 0 0.359871\\
  3 0 0.231291\\
  0 1 0.801836\\
  1 1 0.772698\\
  2 1 0.580518\\
  3 1 0.310061\\
  0 2 0.291195\\
  1 2 0.421137\\
  2 2 0.274905\\
  3 2 -0.235977\\
  0 3 -0.993809\\
  1 3 -0.771598\\
  2 3 -0.977968\\
  3 3 -1.808680\\
}{0}{1}
\nextgroupplot[title={\gls{dinov2}},yticklabels={},title style={yshift=-5pt},xlabel style={yshift=+5pt},]
\DrawGainTable{
  0 0 1.164059\\
  1 0 1.214904\\
  2 0 1.158677\\
  3 0 1.133850\\
  0 1 2.434554\\
  1 1 2.560884\\
  2 1 2.525898\\
  3 1 2.357452\\
  0 2 3.029780\\
  1 2 3.287411\\
  2 2 3.278612\\
  3 2 2.880359\\
  0 3 2.045171\\
  1 3 2.299695\\
  2 3 2.230318\\
  3 3 1.398816\\
}{1}{2}
\nextgroupplot[title={\gls{siglip2}},yticklabels={}, title style={yshift=-5pt}, xlabel style={yshift=+5pt},]
\DrawGainTable{
  0 0 1.478929\\
  1 0 1.504723\\
  2 0 1.560807\\
  3 0 1.468154\\
  0 1 3.391077\\
  1 1 3.483195\\
  2 1 3.566136\\
  3 1 3.348684\\
  0 2 5.029413\\
  1 2 5.497009\\
  2 2 5.677959\\
  3 2 5.525077\\
  0 3 5.088252\\
  1 3 5.801459\\
  2 3 6.217341\\
  3 3 5.870084\\
}{2}{3}
\nextgroupplot[title={\gls{qwen3vl2b}},yticklabels={},colorbar horizontal,
    title style={yshift=-5pt},
    xlabel style={yshift=+5pt},
    colorbar style={
      at={(-1.7,-0.325)},anchor=north,width=8cm,height=0.18cm,
      xtick={-6,-3,0,3,6},
      xticklabel style={font=\fontsize{7.5}{9}\selectfont},
      tick style={black!50,line width=0.3pt},tick align=outside,
      axis line style={black!30,line width=0.3pt},
      xlabel style={font=\fontsize{12}{13.5}\selectfont,yshift=1pt},
    }]
\DrawGainTable{
  0 0 1.040173\\
  1 0 1.072854\\
  2 0 0.982436\\
  3 0 0.936611\\
  0 1 2.245650\\
  1 1 2.298071\\
  2 1 2.254807\\
  3 1 2.059809\\
  0 2 2.751302\\
  1 2 3.013727\\
  2 2 3.095202\\
  3 2 2.762568\\
  0 3 1.643221\\
  1 3 2.126146\\
  2 3 2.210666\\
  3 3 1.551650\\
}{2}{2}
\end{groupplot}
\node[] (colorbar_xlabel) at (11.8, -1.45) {\fontsize{9.5}{11}\selectfont Mean Top-1 gain (\gls{pp})};
\draw[gray!50, rounded corners=2pt, line width=0.3pt] (1.35, -1.9) rectangle (13.45, -0.9);
\end{tikzpicture}%
}
  \caption{Sensitivity analysis of \gls{promd} hyperparameters $m$, $k$ and $\lambda_{\mathrm{den}}$ for \gls{clip}'s \texttt{ViT-B} backbones averaged across 14 datasets. Displays the Top-1 delta over the zero-shot baseline. %
  }
  \label{fig:promd-cross-benchmark-hyp-sensitivity-clip}
\end{figure}
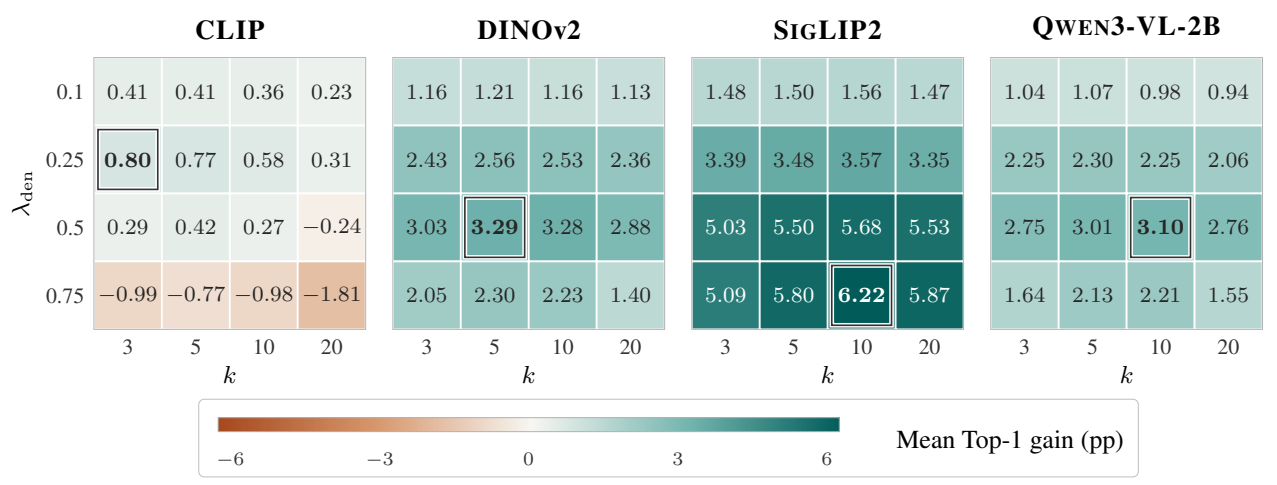

\cref{fig:promd-cross-benchmark-hyp-sensitivity-clip} evaluates \gls{promd}'s denoising hyperparameters $m$, $k$, and $\lambda_{\mathrm{den}}$ for the ViT-B backbones of \gls{clip}. Manifold choice substantially affects the achievable improvement and the preferred $\lambda_{\mathrm{den}}$.\ Using \gls{clip}'s own neighborhoods for denoising consistently leads to the lowest improvement, suggesting it may reinforce the semantic backbone's errors, whereas alternative manifolds can provide complementary visual structure. \gls{siglip2} neighborhoods benefit from stronger interpolation, whereas \gls{clip} neighborhoods degrade predictions at $\lambda_{\mathrm{den}}{=}0.75$, while \gls{dinov2} and \gls{qwen3} achieve the largest gains at $\lambda_{\mathrm{den}}{=}0.5$.\ This illustrates the trade-off characterized by \cref{prop:risk-neighborhood-interpolation}. Increasing $\lambda_{\mathrm{den}}$ strengthens both the corrective contribution of neighborhood consensus and the influence of its errors, making the beneficial range manifold-dependent. Across the tested configurations, performance generally varies less with $k$ than with the manifold or $\lambda_{\mathrm{den}}$.

\subsection{\gls{promd} Gains under Matched Backbone Access}\label{sec:ablations-effect-promd-sota-matched-representations}
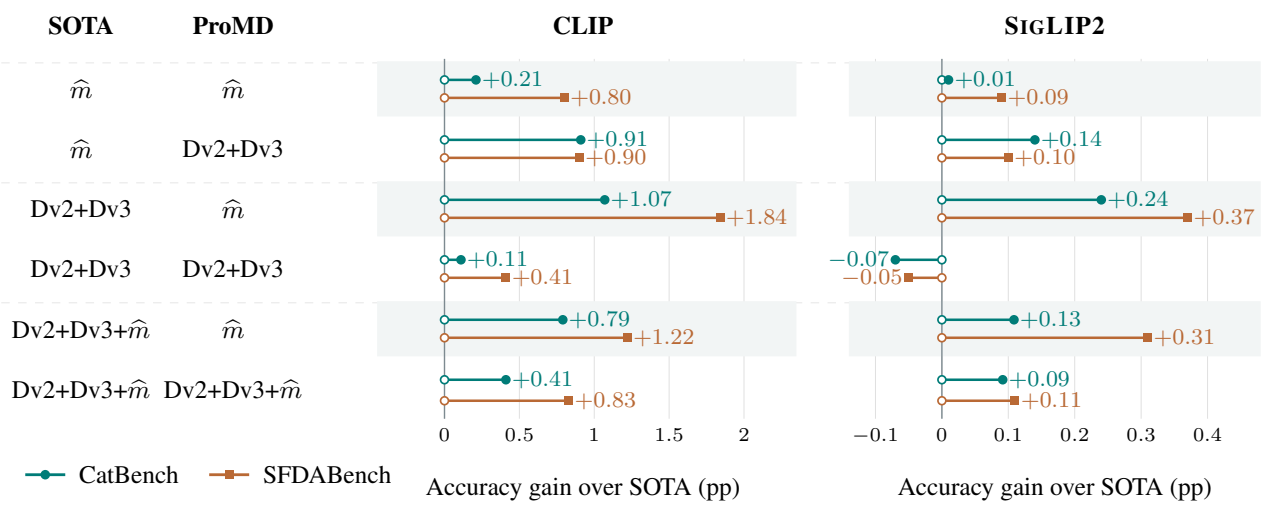
\begin{figure}[t]
\centering
\begingroup
\definecolor{catteal}{HTML}{007C78}
\definecolor{sfdaorange}{HTML}{B96935}
\definecolor{rowshade}{HTML}{F2F5F5}
\definecolor{axisgray}{HTML}{7D8A90}

\newcommand{\gainmark}[5]{%
    \draw[#3,line width=0.9pt]
        (axis cs:0,#2) -- (axis cs:#1,#2);
    \draw[#3,fill=white,line width=0.6pt]
        (axis cs:0,#2) circle (1.4pt);
    \ifnum#4=0
        \fill[#3] (axis cs:#1,#2) circle (1.7pt);
    \else
        \fill[#3] (axis cs:#1,#2)
            ++(-1.6pt,-1.6pt) rectangle ++(3.2pt,3.2pt);
    \fi
    \pgfmathparse{#1 < 0}
    \ifnum\pgfmathresult=1
        \node[anchor=east,text=#3,font=\fontsize{8.5}{9.5}\selectfont,inner sep=2pt]
            at (axis cs:#1,#2) {#5};
    \else
        \node[anchor=west,text=#3,font=\fontsize{8.5}{9.5}\selectfont,inner sep=2pt]
            at (axis cs:#1,#2) {#5};
    \fi
}

\pgfplotsset{
    promdaxis/.style={
        scale only axis,
        width=5.15cm,
        height=4.5cm,
        ymin=-0.5,
        ymax=5.5,
        ytick=\empty,
        axis x line*=bottom,
        axis y line=none,
        axis line style={draw=none},
        tick style={draw=none},
        xticklabel style={font=\scriptsize},
        xmajorgrids=true,
        grid style={draw=black!12,line width=0.25pt},
        xlabel={Accuracy gain over SOTA (pp)},
        xlabel style={
            font=\small,
            at={(axis description cs:0.5,-0.02)},
            anchor=north
        },
        clip=false
    }
}

\adjustbox{max width=\linewidth}{%
\begin{tikzpicture}[font=\small]

\node[font=\small\bfseries] at (1.20,4.92) {SOTA};
\node[font=\small\bfseries] at (3.10,4.92) {ProMD};

\foreach \yy/\src/\dst in {
    4.125/{$\widehat m$}/{$\widehat m$},
    3.375/{$\widehat m$}/{Dv2+Dv3},
    2.625/{Dv2+Dv3}/{$\widehat m$},
    1.875/{Dv2+Dv3}/{Dv2+Dv3},
    1.125/{Dv2+Dv3+$\widehat m$}/{$\widehat m$},
    0.375/{Dv2+Dv3+$\widehat m$}/{Dv2+Dv3+$\widehat m$}
}{
    \node at (1.20,\yy) {\small \src};
    \node at (3.10,\yy) {\small \dst};
}

\draw[catteal,line width=1pt] (0.5,-0.7)--(1,-0.7);
\fill[catteal] (0.75,-0.7) circle (1.7pt);
\node[anchor=west] at (1.05,-0.7) {\small CatBench};

\draw[sfdaorange,line width=1pt] (2.8,-0.7)--(3.3,-0.7);
\fill[sfdaorange] (3.65-0.6,-0.7)
    ++(-1.6pt,-1.6pt) rectangle ++(3.2pt,3.2pt);
\node[anchor=west] at (3.35,-0.7) {\small SFDABench};

\node[font=\small\bfseries] at (7.475,4.92) {CLIP};
\node[font=\small\bfseries] at (13.375,4.92)
    {\gls{siglip2}};

\foreach \yy in {4.5,3.0,1.5}{
    \draw[dashed,gray!20,line width=0.2pt]
        (0.2,\yy-0.05) -- (15.95,\yy-0.05);
}

\begin{axis}[
    promdaxis,
    at={(4.9cm,0cm)},
    anchor=south west,
    xmin=-0.45,
    xmax=2.3,
    xtick={0,0.5,1,1.5,2}
]
    \path[fill=rowshade]
        (axis cs:-0.45,0.54) rectangle (axis cs:2.35,1.46)
        (axis cs:-0.45,2.54) rectangle (axis cs:2.35,3.46)
        (axis cs:-0.45,4.54) rectangle (axis cs:2.35,5.46);

    \draw[axisgray,line width=0.5pt]
        (axis cs:0,-0.5)--(axis cs:0,5.5);

    \gainmark{0.21}{5.15}{catteal}{0}{$+0.21$}
    \gainmark{0.91}{4.15}{catteal}{0}{$+0.91$}
    \gainmark{1.07}{3.15}{catteal}{0}{$+1.07$}
    \gainmark{0.11}{2.15}{catteal}{0}{$+0.11$}
    \gainmark{0.79}{1.15}{catteal}{0}{$+0.79$}
    \gainmark{0.41}{0.15}{catteal}{0}{$+0.41$}

    \gainmark{0.80}{4.85}{sfdaorange}{1}{$+0.80$}
    \gainmark{0.90}{3.85}{sfdaorange}{1}{$+0.90$}
    \gainmark{1.84}{2.85}{sfdaorange}{1}{$+1.84$}
    \gainmark{0.41}{1.85}{sfdaorange}{1}{$+0.41$}
    \gainmark{1.22}{0.85}{sfdaorange}{1}{$+1.22$}
    \gainmark{0.83}{-0.2}{sfdaorange}{1}{$+0.83$}
\end{axis}

\begin{axis}[
    promdaxis,
    at={(10.80cm,0cm)},
    anchor=south west,
    xmin=-0.14,
    xmax=0.48,
    xtick={-0.1,0,0.1,0.2,0.3,0.4}
]
    \path[fill=rowshade]
        (axis cs:-0.14,0.54) rectangle (axis cs:0.48,1.46)
        (axis cs:-0.14,2.54) rectangle (axis cs:0.48,3.46)
        (axis cs:-0.14,4.54) rectangle (axis cs:0.48,5.46);

    \draw[axisgray,line width=0.5pt]
        (axis cs:0,-0.5)--(axis cs:0,5.5);

    \gainmark{0.01}{5.15}{catteal}{0}{$+0.01$}
    \gainmark{0.14}{4.15}{catteal}{0}{$+0.14$}
    \gainmark{0.24}{3.15}{catteal}{0}{$+0.24$}
    \gainmark{-0.07}{2.15}{catteal}{0}{$-0.07$}
    \gainmark{0.108549}{1.15}{catteal}{0}{$+0.13$}
    \gainmark{0.091580}{0.15}{catteal}{0}{$+0.09$}

    \gainmark{0.09}{4.85}{sfdaorange}{1}{$+0.09$}
    \gainmark{0.10}{3.85}{sfdaorange}{1}{$+0.10$}
    \gainmark{0.37}{2.85}{sfdaorange}{1}{$+0.37$}
    \gainmark{-0.05}{1.85}{sfdaorange}{1}{$-0.05$}
    \gainmark{0.31}{0.85}{sfdaorange}{1}{$+0.31$}
    \gainmark{0.11}{-0.2}{sfdaorange}{1}{$+0.11$}
\end{axis}

\end{tikzpicture}%
}
\endgroup

\caption{Effect of ProMD after SOTA under matched access to visual-backbones.\ Dv2, Dv3, and $\widehat m$ denote \gls{dinov2}, \gls{dinov3}, and the selected manifold; gains are relative to each SOTA baseline.}
\label{fig:sota-denoising-effect}
\end{figure}
Since we use published SOTA configuration in \cref{sec:experiments}, we investigate whether the gains of \gls{promd}${+}$ over SOTA arise solely from access to additional visual encoders.\ We use the same prompt scheme and vary the representations available to SOTA and \gls{promd} among \gls{dinov2}+\gls{dinov3}, our selected manifold ($\widehat m$), and their combination.\ When SOTA already accesses all three representations, denoising with $\widehat m$ improves CatBench/SFDABench accuracy by ${+}0.79/{+}1.22$ \gls{pp} with \gls{clip} and ${+}0.13/{+}0.31$ \gls{pp} with \gls{siglip2} (\cref{fig:sota-denoising-effect}). Denoising with all three manifolds also improves both benchmarks, but yields smaller gains than using $\widehat m$ alone.\ The benefit is configuration-dependent: reusing only the DINO manifolds decreases \gls{siglip2} accuracy, whereas denoising with $\widehat m$ improves both benchmarks for all evaluated SOTA configurations. \gls{promd} improves SOTA predictions using representations already available to SOTA, showing that additional encoder access alone does not explain the gains.

\section{Conclusion and Limitations}\label{sec:conclusion-and-limitation}

Motivated by the limited sensitivity of \gls{vlm}-guided \gls{sfda} to source-domain choice, we formulate \gls{sida}, a unifying transductive taxonomy for adaptation without task-specific source artifacts. We advocate separating prediction refinement from target-classifier training and propose \gls{promd}, a training-free adaptation approach for \glspl{vlm} combining proxy-manifold-guided logit denoising with \emph{optional} \gls{lvlm}-based semantic recovery.\ Across both benchmarks, \gls{promd} improves zero-shot predictions, while \gls{promd}${+}$ further improves predictions already refined through transductive adaptation. We also show that the resulting refined pseudo-labels enable effective target-classifier training without task-specific source models.\ Despite eliminating source training, \gls{promd} incurs additional inference costs across pretrained models for proxy-manifold construction and selection, and optional semantic recovery. Finally, identifying residual prediction errors remains a bottleneck for semantic recovery, motivating further investigation into selection strategies.

\section*{LLM Usage Statement}
We used OpenAI’s ChatGPT/GPT models, mainly Astra and Sol, to assist with manuscript preparation, literature search, related-work drafting, code development, and the exploration and formulation of research ideas. Assistance with manuscript preparation included refining author-written text and preparing figures and tables from experimental results. The authors reviewed and verified all incorporated outputs and revised them as needed. No LLMs were used for any data generation, dataset creation, or data annotation. The authors take full responsibility for the manuscript, implementation, experimental design, analysis, and conclusions presented in this work.

{
    \small
    \bibliographystyle{ieeenat_fullname}
    \bibliography{references}
}

\end{document}